\documentclass[10pt,twocolumn,letterpaper]{article}

\usepackage{cvpr}              

\usepackage{graphicx}
\usepackage{amsmath}
\usepackage{amssymb}
\usepackage{booktabs}
\usepackage{bm}
\usepackage{multirow}
\usepackage{diagbox}
\usepackage{float}
\usepackage{fancyhdr}
\usepackage{xcolor}

\usepackage[pagebackref,breaklinks,colorlinks]{hyperref}
\hypersetup{
    colorlinks=true,
    linkcolor=red,
    filecolor=red,      
    urlcolor=red,
}
\usepackage[capitalize]{cleveref}
\crefname{section}{Sec.}{Secs.}
\Crefname{section}{Section}{Sections}
\Crefname{table}{Table}{Tables}
\crefname{table}{Tab.}{Tabs.}

\newcommand{\topcaption}{%
	\setlength{\abovecaptionskip}{2pt}%
	\setlength{\belowcaptionskip}{2pt}%
	\caption}

\def\cvprPaperID{9122} 
\def\confName{CVPR}
\def\confYear{2023}

\begin{document}
	\title{Exploring Structured Semantic Prior \\for Multi Label Recognition with Incomplete Labels}
	
	\author{Zixuan Ding$^{1,4}$\thanks{Equal contributions. $\dagger$ Corresponding author.} \quad Ao Wang$^{2,3,4}$\footnotemark[1] \quad Hui Chen$^{2,3,\dagger}$ \quad Qiang Zhang$^1$ \quad \\ Pengzhang Liu$^5$ \quad Yongjun Bao$^5$ \quad Weipeng Yan$^5$ \quad Jungong Han$^{6,7}$\\
		$^1$Xidian University \quad $^2$Tsinghua University \quad  $^3$BNRist \\ $^4$Hangzhou Zhuoxi Institute of Brain and Intelligence \quad $^5$JD.com \\ $^6$Department of Computer Science, the University of Sheffield, UK \quad \\ $^7$Centre for Machine Intelligence, the University of Sheffield, UK \\
		{\tt\small dingzixuan@stu.xidian.edu.cn \quad wa22@mails.tsinghua.edu.cn \quad qzhang@xidian.edu.cn} \\
		{\tt\small  \{jichenhui2012,jungonghan77\}@gmail.com \quad \{Paul.yan, baoyongjun, liupengzhang\}@jd.com }
}
\maketitle

\begin{abstract}
 Multi-label recognition (MLR) with incomplete labels is very challenging. Recent works strive to explore the image-to-label correspondence in the vision-language model, \ie, CLIP~\cite{radford2021clip}, to compensate for insufficient annotations. In spite of promising performance, they generally overlook the valuable prior about the label-to-label correspondence. In this paper, we advocate remedying the deficiency of label supervision for the MLR with incomplete labels by deriving a structured semantic prior about the label-to-label correspondence via a semantic prior prompter. We then present a novel Semantic Correspondence Prompt Network (SCPNet), which can thoroughly explore the structured semantic prior. A Prior-Enhanced Self-Supervised Learning method is further introduced to enhance the use of the prior. Comprehensive experiments and analyses on several widely used benchmark datasets show that our method significantly outperforms existing methods on all datasets, well demonstrating the effectiveness and the superiority of our method. Our code will be available at \url{https://github.com/jameslahm/SCPNet}.
\end{abstract}

\section{Introduction}
\label{sec:intro}
\begin{figure}[!t] 
	\centering
	\includegraphics[width=0.8\linewidth]{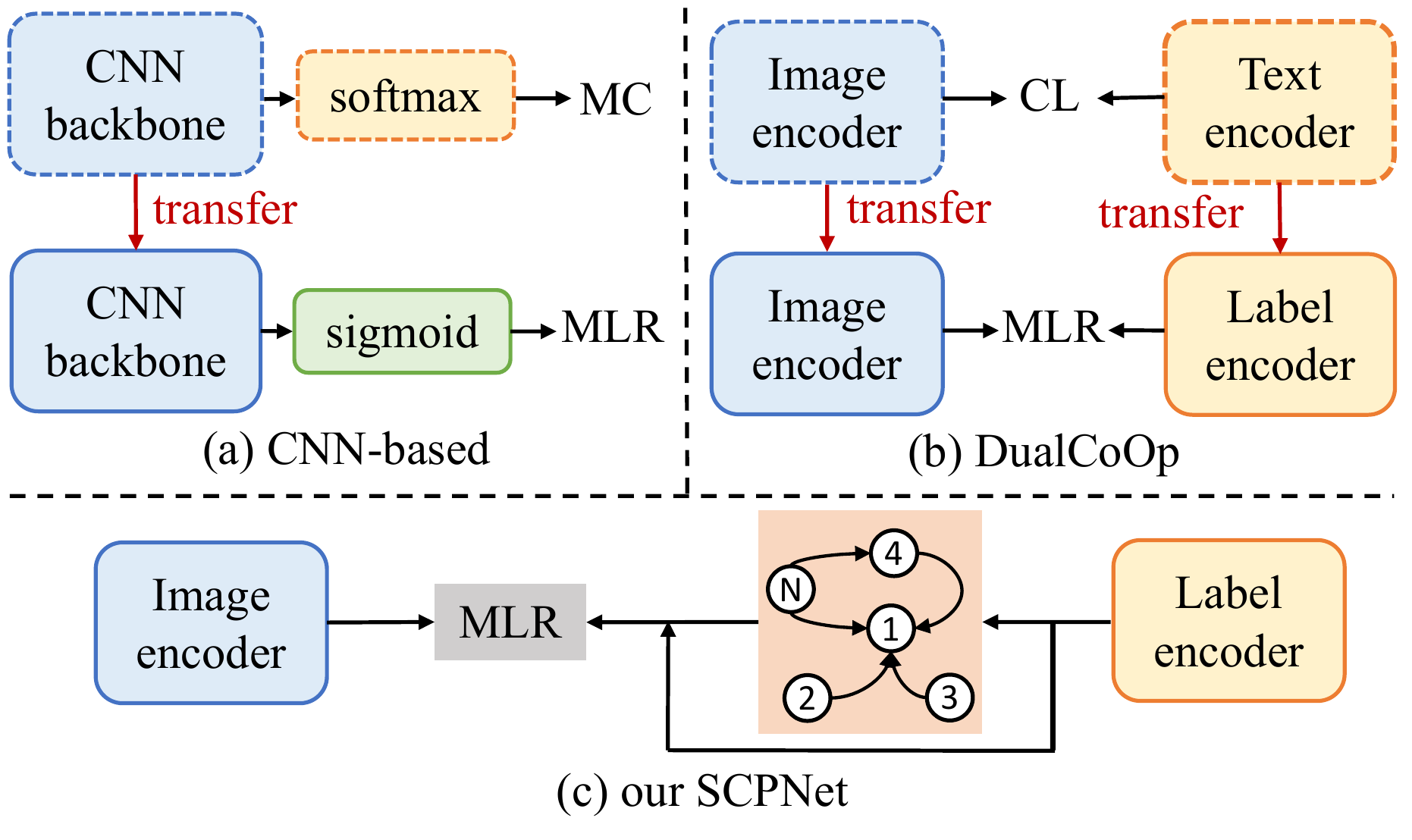}
	\topcaption{Overview of CNN-based, DualCoOp~\cite{sun2022dualcoop} and our SCPNet. Like DualCoOp, our SCPNet adopts CLIP as the base model. Differently, our SCPNet aims to enhance the MLR with the prior about the \textit{label-to-label} correspondence. MC means multi-class. CL denotes contrastive learning.}
	\label{fig:motivation}
\end{figure}
Multi-label recognition (MLR) aims to describe the image content with various semantic labels \cite{yazici2020orderless,wang2020KSSNet,chen2019SSGRL,sun2022dualcoop}. It encodes the visual information into structured labels, which can benefit the index and fast retrieval of images in broad practical applications, such as the search engine \cite{sivic2006videosearch,tautkute2019searchengine} and the recommendation system \cite{carrillo201recommender,zheng2014contextrecommendation}.

Benefited from the development of deep learning, MLR has achieved remarkable progress in recent years. However, collecting high-quality full annotations becomes very challenging when the label set scales up, which greatly hinders the wide usage of MLR in real scenarios. Recently, researchers explore more feasible solutions for MLR. For example, the full label setting is relaxed with a \textit{partial label} setting in \cite{pu2022SARB,chen2020KGGR}, which merely annotates a few labels for each training image. One more extreme setting with solely one \textit{single positive label} is tackled in \cite{kim2022large,cole2021multi}. These settings can be unified into a common issue of \textit{incomplete labels}, which relieves the burden of the full annotation and considerably reduces the annotation cost. Therefore, it draws increasing attention from both academia and industry.

Compared with the full label setting, the incomplete label setting encounters a dilemma of poor supervision, resulting in severe performance drops for MLR. Existing methods strive to regain supervision from missing labels by exhaustively exploring the \textit{image-to-label} correspondence via semantic-aware modules \cite{chen2022SST,pu2022SARB} or loss calibration methods \cite{cole2021multi,zhang2021simple,kim2022large}. A convolutional neural network (CNN) pretrained on the ImageNet is usually leveraged to construct the MLR model. Its multi-class softmax layer is often replaced by a multi-label sigmoid layer (\cref{fig:motivation} (a)). Such a replacement wipes out prior knowledge about the correspondence between images and labels although it is necessary and inevitable.

Recently, vision-language pretrained models have obtained remarkable success in various vision tasks \cite{zhou2022coop,zhou2022cocoop,sun2022dualcoop}. Thanks to their large-scale pretraining, the vision-language model, \eg, CLIP~\cite{radford2021clip}, which is trained with 400 million image-text pairs, can well bridge the visual-textual gap~\cite{sun2022dualcoop}, providing rich prior knowledge for the downstream tasks. For the MLR task, Sun \etal~\cite{sun2022dualcoop} propose a DualCoOp method, which is the first work to employ the CLIP as the MLR base model. Through dual prompts, DualCoOp directly adopts the text encoder in the CLIP as the multi-label classification head (\cref{fig:motivation} (b)), without abandoning the visual-textual prior in the pretrained CLIP.

Despite its effectiveness, DualCoOp is still limited in remedying the deficiency of label supervision, which is desired for the MLR with incomplete labels. Intuitively, it is convenient to reason unknown labels from annotated labels by leveraging the correspondence among labels, \eg, tables are likely to appear with chairs, and cars are usually accompanied by roads. Therefore, such a \textit{label-to-label} correspondence can help survive more label supervision and thus benefit MLR with incomplete labels. Besides, although most vision-language models do not encourage the contrastive learning among texts, they are still abundant in the knowledge about the \textit{label-to-label} correspondence because of the large-scale cross-modality training. However, such a valuable prior is rarely explored in the existing state-of-the-art method, \ie, DualCoOp~\cite{sun2022dualcoop}. 

In this paper, we aim to mitigate such deficiency of label supervision for MLR with incomplete labels by leveraging the abundant prior about the \textit{label-to-label} correspondence in the CLIP~\cite{radford2021clip}. We present a structured prior prompter to conveniently derive a structured semantic prior from the CLIP. Then we propose a novel Semantic Correspondence Prompt network (SCPNet) (\cref{fig:motivation} (c)), which can prompt the structured label-to-label correspondence with a cross-modality prompter. Our SCPNet also equips a semantic association module to explore high-order relationships among labels with the guidance of the derived structured semantic prior. A prior-enhanced self-supervised learning method is further introduced to comprehensively investigate the valuable prior. As a result, our method can neatly calibrate its predicted semantic distribution while maintaining the self-consistency.

To verify the effectiveness of the proposed method for MLR with incomplete labels, we conduct extensive experiments and analyses on a series of widely used benchmark datasets, \ie, MS COCO\cite{lin2014microsoft}, PASCAL VOC\cite{everingham2012voc12}, NUS Wide\cite{chua2009nus}, CUB\cite{wah2011CUB} and OpenImages\cite{krasin2017openimages}. Experimental results show that our method can significantly outperform state-of-the-art methods on all datasets with a maximal improvement of $6.8\%/3.4\%$ mAP for the single positive label setting and the partial label setting, respectively, well demonstrating its effectiveness and superiority.

Overall, our contributions are four folds.
\begin{itemize}
	\item
	We advocate leveraging a structured semantic prior to deal with the deficiency of label supervision for MLR with incomplete labels. To this end, we extract such a prior via a structured prior prompter.
	\item
	We present a semantic correspondence prompt Network (SCPNet) based on a cross-modality prompter and a semantic association module. The SCPNet can adequately explore the structured prior knowledge, thus boosting MLR with incomplete labels. 
	\item
	We design a prior-enhanced self-supervised learning method to further investigate such a structured semantic prior, which can enjoy both distribution refinement and self-consistency.
	\item
	Experimental results show that our method can consistently achieve state-of-the-art performance on all benchmark datasets, revealing the significant effectiveness. Thorough analyses also demonstrate the superiority of our method.
\end{itemize}

\begin{figure*}[t]
	\centering
	\includegraphics[width=15cm]{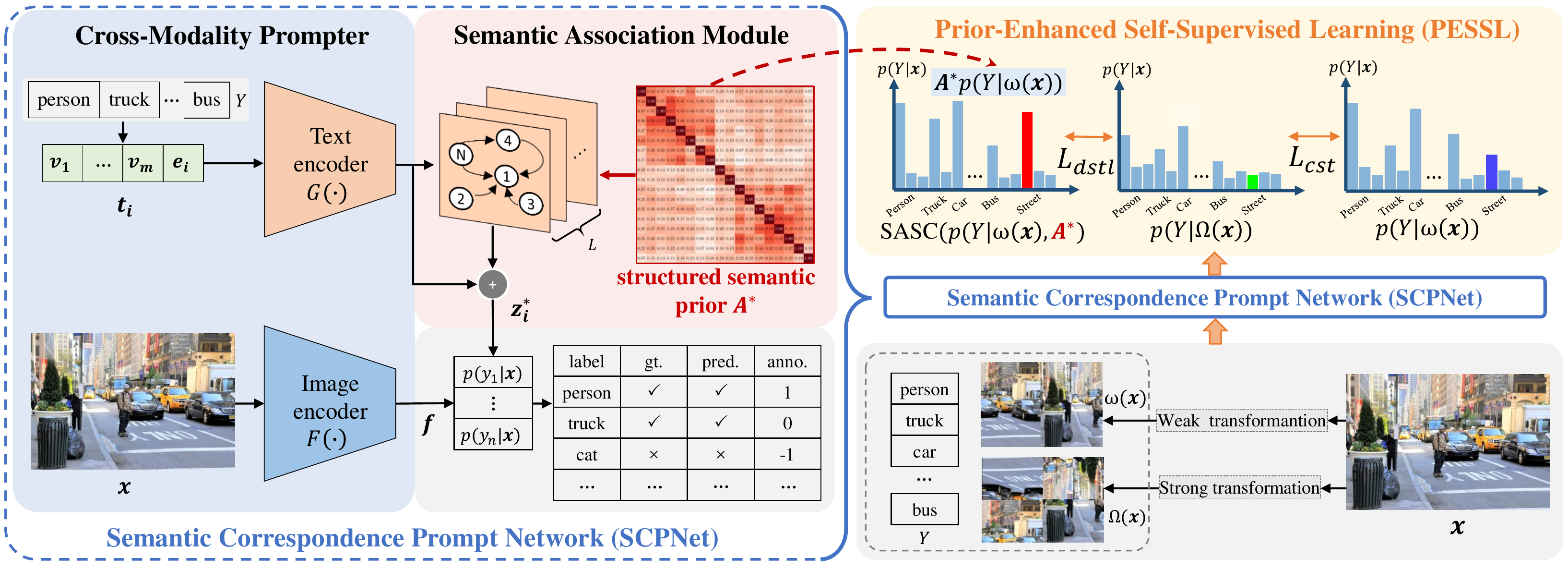}
	\caption{An overview of the proposed method. We design a semantic correspondence prompt network to explore the structured semantic prior for MlR with incomplete labels. A prior-enhanced self-supervised learning strategy is used to enhance such exploration.}
	\label{fig:pipeline}
\end{figure*}

\section{Related work}

\textbf{Multi-label recognition with full annotations.}
Multi-label Recognition has long been a hot topic in the computer vision field~\cite{yazici2020orderless,ben2022P-ASL,pu2022SARB}. A generic method is to learn multiple binary classifiers \cite{cole2021multi,kim2022large}, which usually takes no consideration of the label correlation. Recently, the label-to-label correspondence is established through graph neural networks or transformer structures~\cite{chen2019ML-GCN,wang2020KSSNet}. These methods heavily rely on the quality of label supervision. However, collecting a large-scale dataset with complete labels is challenging and expensive. In real scenarios, researchers explore much more practical settings with incomplete labels, \ie, MLR with partial labels and MLR with a single positive label. 

\textbf{Multi-label recognition with incomplete labels.}
In the partial label setting, only a few labels need to be annotated for each training image. Durand \etal\cite{durand2019learning} adopt a curriculum learning based model to predict the missing labels during the training procedure. Pu \etal \cite{pu2022SARB} and Chen \etal \cite{chen2022SST} transfer predictions of neighboring images via image-image correlation. However, their performance is not guaranteed in more severe scenarios, \ie, single positive label setting, in which each image is provided with solely one positive annotation. To tackle the issue of the single positive label, Cole \etal \cite{cole2021multi} propose a regularized online loss via a joint optimization of label estimator and image classifier. Zhang \etal \cite{zhang2021simple} adopt a label correction process for the probability exceeding a fixed threshold. Kim \etal \cite{kim2022large} propose to reject or correct the large loss samples during training, which can prevent over-fitting false negative labels. However, different from our solution, they usually independently calibrate the importance of different labels~\cite{zhang2021simple,kim2022large,cole2021multi}, taking no consideration of the semantic correspondence among labels. 

\textbf{Vision-language models in downstream visual tasks.}
Radford \etal~\cite{radford2021clip} exploit the contrastive learning with large-scale image-text pairs, \ie, about 400 million pairs, ending up with a powerful vision-language model, \ie, CLIP. Such a model shows remarkable generalization capability in downstream visual tasks~\cite{radford2021clip}. Therefore, researchers exhaustively explore how to leverage the abundant vision-language correspondence~\cite{shin2020autoprompt,gao2020promptfinetune,jiang2020promptcan,sun2022dualcoop}. Sun \etal~\cite{sun2022dualcoop} also employ CLIP for MLR. They present dual prompts, \ie, a positive prompt and a negative one, to explore the rich image-to-label correspondence in CLIP. However, different from our motivation, they overlook the rich label-to-label correspondence in CLIP.

\section{Methodology}

\subsection{Structured Prior Prompter}
\label{sec:prior_extract}

For MLR with full annotations, existing methods can achieve fruitful outcomes by exploring the semantic correspondence between images and labels \cite{chen2019ML-GCN}. However, they require abundant label supervision to obtain accurate label co-occurrence information for the estimation of label relationships. Therefore, in MLR with incomplete labels, the scarce label supervision greatly hinders their capability to explore the semantic correspondence. Benefited from the development of large-scale pretrained embeddings, \eg, Glove~\cite{pennington2014glove}, or models, \eg, BERT~\cite{kenton2019bert} and CLIP~\cite{radford2021clip}, we can easily obtain contextual representations for labels, which can be directly used to derive such a label-to-label correspondence. Such a annotation-free strategy is notably appealing when no adequate label supervision is provided. Furthermore, the abundant correspondence prior in the pretrained model can help associate the annotated label with unknown labels, which promisingly alleviates the deficiency of label supervision. Hence, we introduce a structured prior prompter to explore such a label-to-label correspondence in the pretrained model. Considering the popularity and the remarkable performance in the computer vision community, we choose the vision-language model, \ie, CLIP~\cite{radford2021clip}, as the target.

Specifically, in the proposed structured prior prompter, for a set of to-be-explored labels $Y=\{y_0,y_1,...,y_n\}$, we derive the label feature by feeding a prompt template, \ie, \textit{a photo of a [CLS]}, into the text encoder of CLIP. We denote the label feature as $\bm{\bar{z}_i}$ for each $y_i$. Then the correlation prior among labels, denoted as $\bm{A}=(a_{ij})_{n \times n}$, can be derived as:
\begin{equation}
	a_{ij} = \text{sim}(\bm{\bar{z}_i},\bm{\bar{z}_j})
	\label{equ:a_ij}
\end{equation}
where $\text{sim}(\cdot,\cdot)$ is the cosine similarity.

For each entry $\bm{a}_{i}$, we select the top $K$ elements and set the rest to zero, ending up with a sparse matrix, $\bm{A'}=(a'_{ij})_{n \times n}$:
\begin{equation}
	a'_{ij}= \left\{ \begin{array}{l}
		a_{ij},\quad {\rm{  if }}\ j \in \text{topK}(\bm{a}_{i})\\
		0,\quad\ \ \, {\rm{  if }}\ j \notin \text{topK}(\bm{a}_{i})
	\end{array} \right.
	\label{equ:static_sam}
\end{equation}

Following~\cite{chen2019ML-GCN}, we mitigate the over-smoothness of graph representation by adjusting the sparse graph $\bm{A'}$ as follows:
\begin{equation}
	\bar{a}_{ij}= \left\{ \begin{array}{l}
		( s/\sum\nolimits_{i \ne j'}^n a'_{ij'} ) \times a'_{ij},\quad {\rm{  if }}\ i \ne j\\
		1 - s,\qquad\qquad\qquad\quad\ \ \, {\rm{  if }}\ i = j
	\end{array} \right.
\end{equation}
where $s$ is a hyper-parameter which determines weights assigned to a node itself and its neighboring nodes. 
The label correspondence graph $\mathcal{G}$ can be derived as:
\begin{equation}
\bm{a}_{ij}^{*}=\frac{\text{I} [\bar{a}_{ij} \ne 0]\exp (\bar{a}_{ij}/\tau^{'})} {\sum\nolimits_j \text{I} [\bar{a}_{ij} \ne 0]\exp (\bar{a}_{ij}/\tau^{'})}
\label{equ:prior_graph}
\end{equation}
where $\tau^{'}$ controls the distribution smoothness and $\text{I}[\cdot]$ is an indicator function. We denote the the adjacency matrix of $\mathcal{G}$ as $\bm{A^*}=(a^*_{ij})_{n \times n}$.

We see that $\bm{A^*}$ emphasizes the importance of the node itself and weights other nodes according to their relationships (see \cref{equ:a_ij}). Therefore, the fruitful label correspondence can be encoded in such a structured graph, \ie, $\bm{A^*}$, providing rich structured semantic prior for MLR models.

\subsection{Semantic Correspondence Prompt Network}
\label{subsec:scpnet}
As shown in \cref{fig:pipeline}, the SCPNet consists of a cross-modality prompter and a semantic association module.

\textbf{Cross-modality prompter (CMP).}  
Previous works~\cite{kim2022large,chen2022SST,pu2022SARB} usually employ a convolutional neural network pretrained on ImageNet, \eg, ResNet50. During fine-tuning in the downstream MLR tasks, the prior knowledge about the image-to-label correspondence is generally discarded due to the semantic shift, \ie, different label sets between the ImageNet and the MLR benchmark datasets. Differently, we aim to take full use of such an image-label prior during model optimization. Similar to \cite{sun2022dualcoop}, we resolve the problem of semantic shift by a cross-modality prompter, based on a vision-language model, \ie, CLIP~\cite{radford2021clip}.

Formally, following~\cite{zhou2022coop}, given a label set, \ie, $Y=\{y_0,y_1,...,y_n\}$, we introduce $m$ soft prompt tokens to extract its representation. For ease of explanation, we denote the prompt as $\bm{t}_i=\{\bm{v}_1, \bm{v}_2,..., \bm{v}_m, \bm{e}_i\}$, where $\bm{v}$ with a subscript denotes a soft prompt token and $\bm{e}_i$ is the embedding of $y_i$. The label feature of $y_i$, denoted as $\bm{z_i}$, can be derived by the text encoder of CLIP. For an input image $\bm{x}$, its visual representation, denoted as $\bm{f}$, is extracted by the image encoder of CLIP. The process of feature extraction can be computed as follows:
\begin{equation}
	{\bm{f}} = {F}(\bm{x}),{\bm{z_i}} = {G}(\bm{t_i}),
	\label{eq:cmp_feature}
\end{equation}
where ${F(\cdot)}$ and ${G(\cdot)}$ denote the image encoder and the text encoder in CLIP, respectively. 

\textbf{Semantic association module (SAM).} As CMP still lacks capturing the label-to-label correspondence, we further equip a semantic association module to capture high-order relationships among labels. Specifically, with guidance of the structured semantic prior $\bm{A}^{*}$ (see \cref{equ:prior_graph}), we utilize $L$ graph convolutional network (GCN) layers to progressively refine the input features $\bm{H}^0=\bm{Z}$, where $\bm{Z}=\{\bm{z}_0, \bm{z}_1, ..., \bm{z}_n\}$ is a combination of features for $Y$ as in ~\cref{eq:cmp_feature}. The $l$-th GCN layer is updated as follows:
\begin{equation}
	{\bm{H}^{l + 1}} = \rho ({{\bm{A}}^{*}}{\bm{H}^l}{\bm{W}^l}) ,
\end{equation}
where $\bm{W}$ with a superscript is a learnable parameter matrix and $\rho$ is a non-linear function. $l\in [0,L)$. The final refined label representations can be obtained through a residual connection, \ie, $\bm{Z}^{*} = \bm{H}^0 + \bm{H}^L$. The likelihood $p(y_i|\bm{x})$ can be computed as:
\begin{equation}
	p(y_i|\bm{x}) = \sigma (\text{sim} (\bm{f},\bm{z_i}^{*})/\tau ),
	\label{eq:sampro}
\end{equation}
where ${\bm{z_i}}^{*}$ denotes the refined feature for label $y_i$.

Benefited from the GCN, the structured label-to-label correspondence in CLIP, which is represented by $\bm{A}^{*}$, can be progressively refined in the label representation. Therefore, during the semantic matching between the image feature and the label feature, \ie, \cref{eq:sampro}, labels with high correlations will obtain similar likelihoods, enabling a subtle semantic association.

\subsection{Prior-Enhanced Self-Supervised Learning}
\label{subsec:pessl}

The proposed prior-enhanced self-supervised learning strategy, dubbed PESSL, aims to make full use of the structured semantic correspondence prior. We endow the proposed PESSL with a self-supervised consistency loss and a self-distillation objective that is boosted by a structure-aware semantic calibration strategy.

\textbf{Structure-aware semantic calibration.} Intuitively, if two labels are semantically correlated, they may be observed in one image. For MLR, such a correspondence can help decide potential semantic labels for an input image, given its predictions. Therefore, we formulate the likelihood of $p(y_i | \bm{x})$ as a weighted combination of likelihoods for correlated neighboring labels of $y_i$:

\begin{equation}
	p^*(y_i|\bm{x}) = \sum\nolimits_{y_j \in \mathcal N (y_i)} w(i,j) \times p(y_j | \bm{x})
\end{equation} 
Here, $w(i,j)$ is a correlation weight indicating the relationship between $y_i$ and $y_j$. $\mathcal N (y_i)$ denotes a correlated neighboring set of labels corresponding to $y_i$.

For ease of explanation, we introduce a correlation matrix $\mathbb{W}$ to represent the whole correlation among labels, \ie, $\mathbb{W}=(w(i,j))_{n \times n}$. We then customize the whole process as a function parameterized by $\mathbb{W}\in R^{n \times n}$ and the distribution over $Y$ given the input $\bm{x}$, \ie, $\bm{p}(Y|\bm{x}) \in R^{n\times 1}$:
\begin{equation}
	\text{SASC}(\bm{p}(Y|\bm{x}), \mathbb{W})=\mathbb{W}\bm{p}(Y|\bm{x})
	\label{eq:sasc}
\end{equation}

\textbf{Prior-enhanced learning.} Existing loss correction methods individually reweight each label, without taking into consideration the correspondence among labels. Here, we propose to follow the self-supervised learning principle~\cite{zhang2021flexmatch,sohn2020fixmatch} and introduce a self-distillation learning strategy to benefit the MLR model from the structured semantic correspondence among labels.

Specifically, we derive two different versions for the input image $\bm{x}$ with one weak transformation $\omega(\cdot)$ and one strong transformation $\Omega(\cdot)$, respectively. Their corresponding semantic distributions, denoted as $\bm{p}(y|\omega (\bm{x}))$ and $\bm{p}(y|\Omega (\bm{x}))$, respectively, can be derived by \cref{eq:sampro}. Then we use a consistency loss to encourage them to be consistent. Different from ~\cite{zhang2021flexmatch}, which simply regularizes the model with the most confident label, we construct a set of confident labels $\mathcal{O}(\bm{x})$ with the top highest probability larger than a threshold $\mathcal{T}$ in $\bm{p}(y|\omega (\bm{x}))$, \ie, $\mathcal{O}(\bm{x})=\{c | c \in \text{topK}(\bm{p}(y|\omega (\bm{x}))) \wedge p(c|\omega (\bm{x})) > \mathcal{T}(c)\}$. A dynamic threshold strategy is performed for each label, as~\cite{zhang2021flexmatch}. 
The consistency loss is then derived by:
\begin{equation}
	\begin{aligned}
		\mathcal{L}_{cst} = &-\sum\nolimits_{c \in \mathcal{O}(\bm{x})}^ Y \log p(c|\Omega (\bm{x})) \\ &-\sum\nolimits_{c \notin \mathcal{O}(\bm{x})}^ Y \log (1-p(c|\Omega (\bm{x})))
	\end{aligned}
\end{equation}

We calibrate the distribution of the weak-transformed image, \ie, $\bm{p}(y|\omega (\bm{x}))$, by using the SASC function (see ~\cref{eq:sasc}):
\begin{equation}
	\bm{p^*}(y|\omega (\bm{x})) = \text{SASC}(\bm{p}(y|\omega (\bm{x})), \bm{A}^{*})
\end{equation}
where $\bm{A}^{*}$ represents the structured semantic prior, derived by \cref{equ:prior_graph}. 
Considering that compared with the weak-transformed image, the strong-transformed image is usually more difficult to learn. Therefore, we employ a self-distillation objective to optimize the distribution of the strong-transformed image $\Omega (\bm{x})$ with the guidance of the calibrated semantic distribution via the KL-divergence:
\begin{equation}
	\mathcal{L}_{dstl} = -\sum\limits_c^Y  \left( q^w_c \log \frac{q_c^s}{q^w_c} + {{(1- q^w_c)} \log \frac{1 - {q^s_c}}{1- q^w_c}} \right)
\end{equation}
where $q^w_c = p^*(c|\omega (\bm{x}))$ and $q^s_c=p(c|\Omega (\bm{x}))$.

\textbf{Overall Objective.}
Finally, we formulate the prior-enhanced self-supervised learning as a combination of the consistency objective and the self-distillation objective:
\begin{equation}
	\mathcal{L}_{pessl} = \lambda_{cst} \mathcal{L}_{cst} + \lambda_{dstl} \mathcal{L}_{dstl}
\end{equation}

\subsection{Network Optimization} 

During training, we adopt a multi-label classification objective over the predicted likelihood, \ie, $p(y_i|\bm{x})$ in \cref{eq:sampro}, to optimize our SCPNet, denoted as $\mathcal{L}_{cls}$. We follow \cite{zhang2021simple} to design $\mathcal{L}_{cls}$. The overall objective for the network optimization is formulated as follows:
\begin{equation}
\mathcal{L} = \mathcal{L}_{cls} + \mathcal{L}_{pessl}
\end{equation}

\begin{table*}[htbp]
\centering
\caption{Comparison with the state-of-the-art methods for MLR with the single positive label (\%).}
\begin{tabular}{c|ccccc|ccccc}
\hline
\multirow{2}{*}{Method}      & \multicolumn{5}{c|}{LargeLoss setup~\cite{kim2022large}}             & \multicolumn{5}{c}{SPLC setup~\cite{zhang2021simple}}                  \\
\cline{2-11}
                             & COCO          & VOC           & NUS           & CUB          &Avg.   & COCO          & VOC           & NUS     & CUB   &Avg.           \\ \hline
LSAN\cite{cole2021multi}     & 69.2          & 86.7          & 50.5          & 17.9         &56.1   & 70.5          & 87.2          & 52.5    & 18.9  & 57.3   \\
ROLE\cite{cole2021multi}     & 69.0          & 88.2          &{51.0}  & 16.8      &56.3   & 70.9          & {89.0} & 50.6 & {20.4}  & 57.7 \\
LargeLoss\cite{kim2022large} & 71.6          & {89.3}  & 49.6  & {21.8} &{58.1} & -   & -             & -       & -     & -  \\
Hill\cite{zhang2021simple}   & -             & -             & -             & -            & -     & 73.2          & 87.8          & 55.0    & 18.8  & 58.7  \\
SPLC\cite{zhang2021simple}   & {72.0} & 87.7       & 49.8          & 18.0         & 56.9  & {73.2} & 88.1 & {55.2} & 20.0 & {59.1}   \\  \hline
\textbf{SCPNet (ours)}       & \textbf{75.4} & \textbf{90.1} & \textbf{55.7} & \textbf{25.4} &\textbf{61.7} &\textbf{76.4} & \textbf{91.2} & \textbf{62.0} & \textbf{25.7} & \textbf{63.8} \\ 
\hline
\end{tabular}

\label{tab:single}
\end{table*}

\begin{table*}[htbp]
\centering
\caption{Comparison with the state-of-the-art methods for MLR with partial labels (\%).}
\begin{tabular}{c|c|ccccccccc|c}
\hline
Datasets                & Method                               & 10\%      & 20\%      & 30\%      & 40\%      & 50\%      & 60\%      & 70\%      & 80\%      & 90\%      & Avg.       \\ \hline
\multirow{7}{*}{COCO}   & SSGRL\cite{chen2019SSGRL}            & 62.5      & 70.5      & 73.2      & 74.5      & 76.3      & 76.5      & 77.1      & 77.9      & 78.4      & 74.1           \\
                        & GCN-ML\cite{chen2019ML-GCN}          & 63.8      & 70.9      & 72.8      & 74.0      & 76.7      & 77.1      & 77.3      & 78.3      & 78.6      & 74.4           \\
                        & SST\cite{chen2022SST}                & 68.1      & 73.5      & 75.9      & 77.3      & 78.1      & 78.9      & 79.2      & 79.6      & 79.9      & 76.7           \\
                        & SARB\cite{pu2022SARB}                & 71.2      & 75.0      & 77.1      & 78.3      & 78.9      & 79.6      & 79.8      & 80.5      & 80.5      & 77.9           \\
                        & DualCoOp\cite{sun2022dualcoop}       & 78.7      & 80.9      & 81.7      & 82.0      & 82.5      & 82.7      & 82.8      & 83.0      & 83.1      & 81.9           \\                 
                        & \textbf{SCPNet (ours)*}              & \textbf{80.3}      & \textbf{82.2}      & \textbf{82.8}     & 83.4      & 83.8      & 83.9      & 84.0      & 84.1      & 84.2      &83.2 \\
                        & \textbf{SCPNet (ours)}               & 79.1  & 82.1 & \textbf{82.8} & \textbf{83.9} & \textbf{84.5} & \textbf{84.9} & \textbf{85.4} & \textbf{85.7} & \textbf{85.9} & \textbf{83.8} \\ \hline
\multirow{6}{*}{VOC2007}& SSGRL\cite{chen2019SSGRL}            & 77.7      & 87.6      & 89.9      & 90.7      & 91.4      & 91.8      & 91.9      & 92.2      & 92.2      & 89.5           \\
                        & GCN-ML\cite{chen2019ML-GCN}          & 74.5      & 87.4      & 89.7      & 90.7      & 91.0      & 91.3      & 91.5      & 91.8      & 92.0      & 88.9           \\
                        & SST\cite{chen2022SST}                & 81.5      & 89.0      & 90.3      & 91.0      & 91.6      & 92.0      & 92.5      & 92.6      & 92.7      & 90.4           \\
                        & SARB\cite{pu2022SARB}                & 83.5      & 88.6      & 90.7      & 91.4      & 91.9      & 92.2      & 92.6      & 92.8      & 92.9      & 90.7           \\
                        & DualCoOp\cite{sun2022dualcoop}       & {90.3}  & {92.2} & {92.8} & {93.3} & {93.6} & {93.9} & {94.0} & {94.1} & {94.2} & {93.2}           \\
                        & \textbf{SCPNet (ours)}               & \textbf{91.1}     & \textbf{92.8}    & \textbf{93.5}    & \textbf{93.6}    & \textbf{93.8}    & \textbf{94.0}    & \textbf{94.1}    & \textbf{94.2}    & \textbf{94.3}    & \textbf{93.5}           \\ 
                        \hline
\multirow{5}{*}{VG-200} & SSGRL\cite{chen2019SSGRL}            & 34.6      & 37.3      & 39.2      & 40.1      & 40.4      & 41.0      & 41.3      & 41.6      & 42.1      & 39.7           \\
                        & GCN-ML\cite{chen2019ML-GCN}          & 32.0      & 37.8      & 38.8      & 39.1      & 39.6      & 40.0      & 41.9      & 42.3      & 42.5      & 39.3           \\
                        & SST\cite{chen2022SST}                & 38.8      & 39.4      & 41.1      & 41.8      & 42.7      & 42.9      & 43.0      & 43.2      & 43.5      & 41.8           \\
                        & SARB\cite{pu2022SARB}                & {41.4} & {44.0} & {44.8} & {45.5} &{46.6} & {47.5} & {47.8} & {48.0} & {48.2} & {46.0}           \\
                        & \textbf{SCPNet (ours)}               & \textbf{43.8} & \textbf{46.4} & \textbf{48.2} & \textbf{49.6} & \textbf{50.4} & \textbf{50.9} & \textbf{51.3} & \textbf{51.6} & \textbf{52.0} & \textbf{49.4} \\\hline
\end{tabular}
\label{tab:partial}
\end{table*}

\section{Experiment} 

\subsection{Experiment Settings}
\label{subsec:incomplete dataset}

\textbf{Datasets.}
We conduct extensive experiments on several standard benchmarks for MLR with incomplete labels, including the single positive label setting and the partial label setting. For the single positive label setting, following \cite{kim2022large,zhang2021simple}, we use MS-COCO (COCO)~\cite{lin2014microsoft}, PASCAL VOC (VOC)~\cite{everingham2012voc12}, NUSWIDE (NUS)~\cite{chua2009nus}, and CUB~\cite{wah2011CUB}. For the partial label learning, we adopt MS-COCO (COCO)~\cite{lin2014microsoft}, PASCAL VOC 2007 (VOC2007)~\cite{everingham2010voc07} and Visual Genome (VG-200)~\cite{krishna2017VG}, as \cite{chen2022SST,pu2022SARB}. We leave details of benchmark datasets in the supplementary due to the space limit.

\textbf{Implementation details.}
We leverage published CLIP weights\footnote{https://github.com/openai/CLIP} to initialize MLR models. To fairly compare the proposed method with others, we adopt the ResNet50-based CLIP and the Resnet101-based CLIP for the single positive label and the partial label, respectively. During training, we tune the image encoder and fix the text encoder of CLIP. More details are provided in the supplementary.

\textbf{Evaluation.} By default, we employ the mean average precision (mAP) as the evaluation metric, following previous works~\cite{chen2019SSGRL,zhang2021simple,kim2022large}.
For the single positive label setting, we perform two different setups, \ie, the LargeLoss setup \cite{kim2022large} and the SPLC setup~\cite{zhang2021simple}, which are common in the community. We leave the details in the supplementary due to the space limit. For the partial label setting, following \cite{pu2022SARB}, we randomly maintain partial labels for the training set with a ratio ranging from 10$\%$ to 90$\%$. Apart from performance on all ratios, we also report the average result. 

\begin{table*}[htbp]
	\centering
	\caption{Effect of different modules in the proposed SCPNet method for both the single positive label setting and the partial label setting (\%). An average of all metrics is also reported.}
		\renewcommand{\arraystretch}{1.2}
		\begin{tabular}{c|cc|cc|cccc|ccc|c}
			\hline
			\multirow{2}{*}{Model} & \multirow{2}{*}{CMP} & \multirow{2}{*}{SAM} & \multicolumn{2}{c|}{PESSL} & \multicolumn{4}{c|}{Single Positive Label} & \multicolumn{3}{c|}{Partial Label} &\multirow{2}{*}{Avg.}     \\ \cline{4-12} 
			                       &           &  & $\mathcal{L}_{cst}$ & $\mathcal{L}_{dstl}$  & COCO   & VOC            & NUS            & CUB            & COCO           & VOC2007        & VG-200          \\ \hline
			Baseline               &              &              &             &              & 73.18          & 88.07          & 55.18          & 19.99          & 77.41          & 88.32          & 46.39          & 64.08 \\ \hline
			\multirow{5}{*}{SCPNet}& \checkmark   &              &             &              & 74.36          & 88.46          & 60.66          & 21.42          & 80.90          & 89.16          & 47.55          & 66.07 \\
			                       & \checkmark   & \checkmark   &             &              & 75.12          & 89.09          & 61.08          & 21.66          & 82.12          & 90.16          & 48.11          & 66.76 \\
			                       & \checkmark   & \checkmark   & \checkmark  &              & 75.70          & 90.92          & 61.75          & 23.67          & 82.85          & 92.50          & 48.70          & 68.01 \\
			                       & \checkmark   & \checkmark   &             & \checkmark   & 75.84          & 90.92          & 61.56          & 24.51          & 83.35          & 93.21          & 48.83          & 68.32 \\
			                       & \checkmark   & \checkmark   & \checkmark  & \checkmark   & \textbf{76.42} & \textbf{91.16} & \textbf{62.04} & \textbf{25.71} & \textbf{83.76} & \textbf{93.49} & \textbf{49.36} & \textbf{68.85} \\ \hline
		\end{tabular}
		\label{tab:module ablation}
\end{table*}

\begin{table}[]
	\centering
	\renewcommand{\arraystretch}{1.2}
	\caption{Analysis on the correlation graph.}
	\begin{tabular}{c|c|c}
		\hline
		SAM                        & PESSL        & mAP (\%)       \\ \hline
		\multirow{3}{*}{Static}    & Static       & \textbf{76.42} \\
		                           & Dynamic      & 76.05          \\
		                           & No      & 75.83               \\ \hline
		\multirow{2}{*}{Dynamic}   & Static       & 76.08          \\
		                           & Dynamic      & 75.84          \\ \hline
	\end{tabular}
	\label{tab:correlation_graph}
\end{table}

\subsection{Comparisons with State-of-the-Arts}

\textbf{MLR with single positive labels.} We report the model performance on both the LargeLoss setup~\cite{kim2022large} and the SPLC setup~\cite{zhang2021simple}. To better reveal the effectiveness of the proposed method, we also report the average performance for both setups. As shown in \cref{tab:single}, for both setups, our method can significantly outperform existing methods on all benchmark datasets, achieving state-of-the-art performance. Specifically, in the LargeLoss setup, the proposed SCPNet can obtain a maximal performance improvement of $4.7\%$ (NUS). As a whole, our method can accomplish an overall performance improvement of $3.6\%$. In the SPLC setup, the maximal performance improvement achieved by our method can reach $6.8\%$ (NUS). As a result, our method can accomplish $4.7\%$ improvement on average.

\textbf{MLR with partial labels.}
As shown in \cref{tab:partial}, our results also consistently surpass existing state-of-the-art methods on all benchmark datasets, especially on the COCO and VG-200.  Compared with DualCoOp~\cite{sun2022dualcoop} which also leverages CLIP to build MLR models, the proposed method can obtain an improvement of $1.9\%$ mAP on the MS COCO. With a frozen image encoder during training as DualCoOp, our method, denoted as SCPNet (ours)*, still enjoys superior performance to DualCoOp. On the VOC2007, our method obtains comparable performance with $0.3\%$ improvement. However, under small ratios, our method shows its superiority to DualCoOp, \eg, $0.8\%$ improvement with a ratio of $10\%$. On the VG-200, compared with SARB~\cite{pu2022SARB} which enhances the MLR models with a structure-aware algorithm, our SCPNet can significantly outperform it with an average performance improvement of $3.4\%$. 

These experimental results show that our method can consistently obtain superior performance in different setups for MLR with incomplete labels, well demonstrating the effectiveness. To verify the generalization of the proposed method, we also investigate the effectiveness in the few-shot partial label setting and the real partial label scenario. We leave them in the supplementary due to the space limit.

\subsection{Ablation Study}
In order to analyze the effectiveness of each component, we conduct the ablation study on both the single positive label and the partial label settings. All results are shown in \cref{tab:module ablation}. We also introduce a model that directly employs $\mathcal{L}_{cls}$ to optimize a ResNet-based MLR model, as the baseline. As shown in \cref{tab:module ablation}, each component can obtain consistent performance improvement in all datasets. Specifically, compared with the baseline model, our CMP can obtain an average performance of $1.99\%$ mAP, indicating the superiority of prompting a cross-modality vision-language model. Augmented by SAM, our method can bring $0.69\%$ mAP improvement. Such improvements can be attributed to the explicit semantic correspondence among labels captured by the proposed SAM component. Besides, the consistency learning, \ie, $\mathcal{L}_{cst}$, and the self-distillation objective, \ie, $\mathcal{L}_{dstl}$, can lead to $1.25\%$ and $1.56\%$ performance improvement, respectively. The overall improvement for the proposed PESSL can reach $2.09\%$, well demonstrating the strength of incorporating the structured semantic prior during model optimization. Finally, our proposed SCPNet can significantly outperform the baseline model with $4.77\%$ mAP improvement on average, well demonstrating the effectiveness and the superiority of the proposed method.

\begin{table}[]
	\centering
	\caption{Analysis on the prior extraction (\%).}
	\renewcommand{\arraystretch}{1.2}
	\begin{tabular}{l|lllll}
		\hline
		Prior     & Dynamic &Image & Glove & BERT & CLIP          \\ \hline
		mAP  & 75.84 &75.67 & 76.15 & 76.16 & \textbf{76.42} \\ \hline
	\end{tabular}
	\label{tab:prior_construction}
\end{table}

\subsection{Model Analysis}
Here, we perform comprehensive inspections for the proposed method. All experiments are conducted in the single positive label setting on the MS COCO dataset, by default. Due to the space limit, we provide more analyses in the supplementary material.

\begin{figure*}[htbp]
	\centering
	\begin{subfigure}{.3\linewidth}
		\centering
		\includegraphics[scale=0.23]{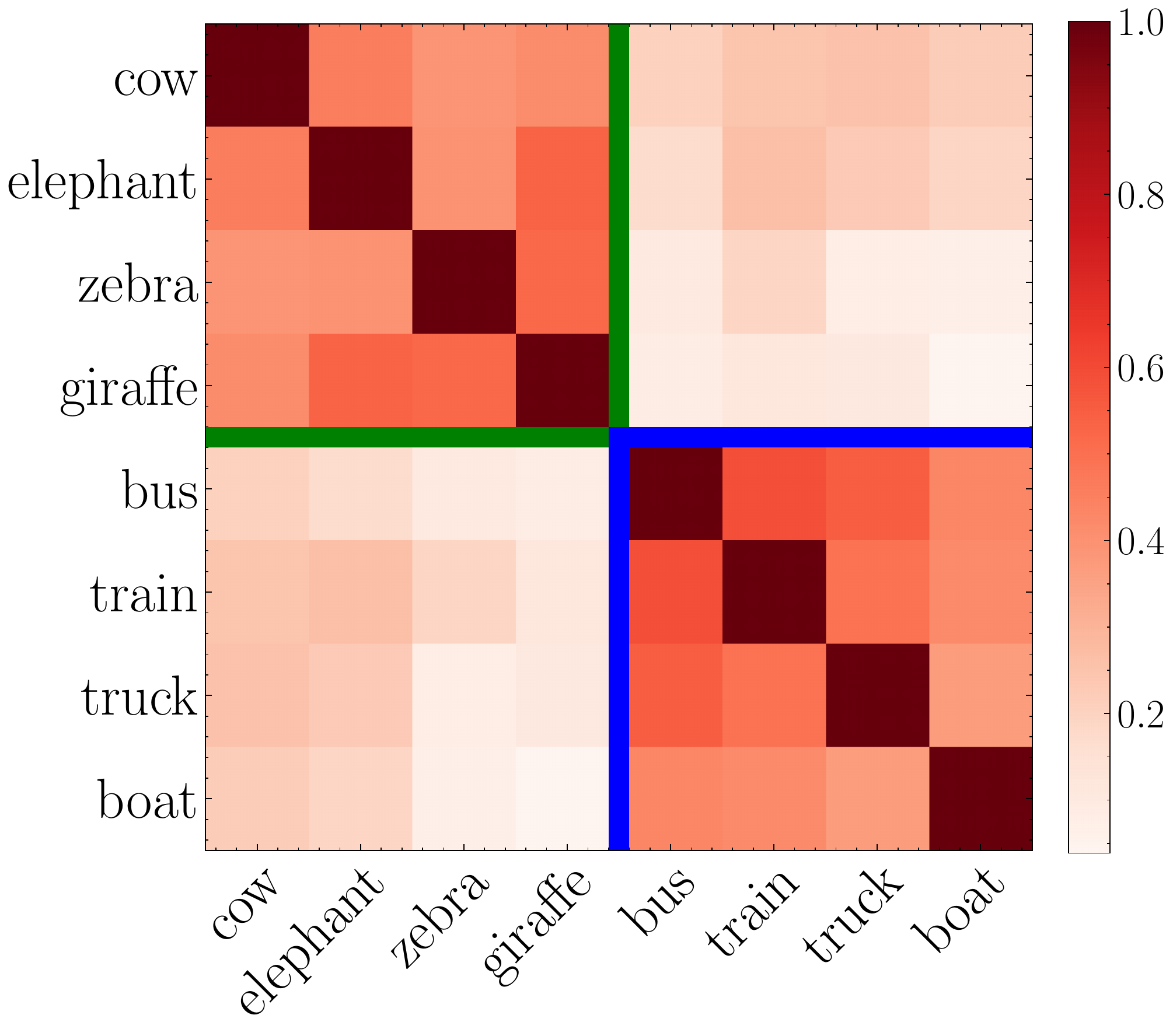}
	\end{subfigure}
	\begin{subfigure}{.68\linewidth}
		\centering
		\includegraphics[scale=0.2]{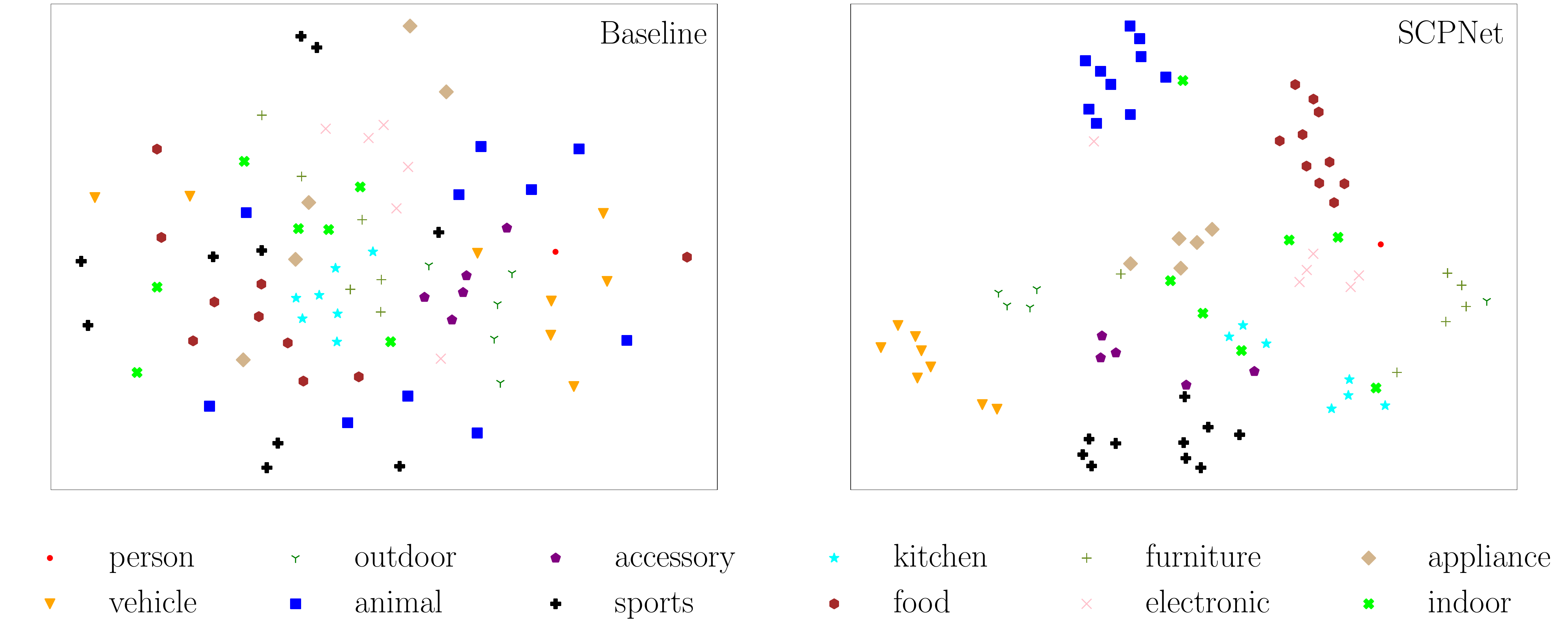}
	\end{subfigure}
	\caption{The structured semantic prior (left) and the learnt label representation (middle: in the baseline, right: in our SCPNet).}
	\label{fig:vis}
\end{figure*}

\textbf{Correlation graph construction.} We verify the positive effect of the prior used in the correlation graph construction for both SAM and PESSL. To achieve this goal, we discuss two kinds of correlation graph: 1) a static one derived from the pretrained CLIP model (see \cref{equ:prior_graph}), which captures the structured semantic prior, and 2) a dynamic one achieved by the learnable CMP, \ie, constructing the adjacency matrix with label features ${\bm{z_i}}$ computed by \cref{eq:cmp_feature}. We also report PESSL without the prior, denoted as ``No''. As illustrated in \cref{tab:correlation_graph}, we can observe that our method can obtain the optimal performance by using the static correlation graph for both SAM and PESSL. Besides, the static graph can substantially achieve better results than the dynamic one in both components, revealing the advantage of the structured semantic prior. We claim that in the MLR with incomplete labels, the challenge of insufficient label supervision makes the dynamic graph sub-optimal, thus inferior to the static one. By comparing PESSL with the prior (Row 2) and the one without the prior (Row 4), we can find that the latter achieves inferior performance, which can demonstrate the benefit of the proposed prior, \ie, $\bm{A}^{*}$ in \cref{equ:prior_graph}.

\textbf{Prior knowledge extraction.} We further investigate the advantage of the proposed structured semantic prior extracted by CLIP with three other types of prior knowledge as competitors. For a given label, 1) ``Image'' averages all image features corresponding to it; 2) ``Glove'' represents its feature by pretrained Glove word embeddings~\cite{pennington2014glove}; and 3) ``BERT'' extracts the label feature by the prompt learning as ours. We also report the result of dynamic label-to-label correspondence as the baseline. As shown in \cref{tab:prior_construction}, compared with Dynamic, except Image, Glove, BERT and CLIP show consistent advantage because of their superior ability to capture the label-to-label correspondence. Besides, our CLIP achieves the best performance, which reveals that the CLIP-based structured prior is more matched with the CLIP-based MLR model due to their consistent knowledge.

\begin{table}[]
	\centering
	\renewcommand{\arraystretch}{1.2}
	\caption{Prior for MLR models with the ImageNet-based ResNet.}
	\begin{tabular}{c|c|c}
		\hline
		Image Encoder          & Label Encoder  & mAP (\%)    \\ \hline
		ResNet  & sigmoid & 73.18 \\
		ResNet & Ours & 74.72 \\
		Ours & Ours & \textbf{76.42} \\
		\hline
	\end{tabular}
	\label{tab:imagenet-resnet}
\end{table}

\textbf{Generalization on the CNN-based architecture.} To show the generalization of the proposed prior-enhanced method, we transfer our design principles to a vanilla ResNet-based MLR model with a sigmoid layer as the label encoder. We analyze the impact of replacing the sigmoid layer with ours. As shown in \cref{tab:imagenet-resnet}, such modification can result in a performance gain of $1.54\%$, demonstrating the good generalization ability of the proposed method in the CNN-based architecture.

\textbf{Analysis on hyper-parameters.} As shown in \cref{tab:gcn_layer} and \cref{tab:parameter}, the best value of $L$, $\lambda_{cst}$ and $\lambda_{dstl}$ is at $L=3$, $\lambda_{cst}=1/8$, and $\lambda_{dstl}=1/8$, respectively. More analyses can be found in the supplementary material.

\begin{table}[]
	\centering
	\caption{Analysis on the number of GCN Layer, \ie, $L$ (\%).}
	\renewcommand{\arraystretch}{1.2}
	\begin{tabular}{c|c|c|c}
		\hline
		$L$ & 2   &3          &4  \\ \hline
		mAP & 75.88    &\textbf{76.42}  &76.22  \\ \hline
	\end{tabular}
	\label{tab:gcn_layer}
\end{table}

\begin{table}[]
	\centering
	\caption{Analysis on $\lambda_{cst}$ and $\lambda_{dstl}$ (\%).}
	\setlength{\tabcolsep}{1mm}{
		\renewcommand{\arraystretch}{1.2}
		\begin{tabular}{c|c|c|c|c|c|c|c}
			\hline
			$\lambda_{cst}$  & 0 & 1/16  & 1/8  & 1/4  & \multicolumn{3}{c}{1/8} \\ \hline
			$\lambda_{dstl}$ & \multicolumn{4}{c|}{0}  & 1/8 & 2/8 & 3/8 \\ \hline
			mAP         &75.12     & 75.56    & \textbf{75.70}   & 75.13     & \textbf{76.42}   & 76.40   & 76.17    \\ \hline
		\end{tabular}
	}
	\label{tab:parameter}
\end{table}

\subsection{More Insightful Analysis}

To provide more insights about the effectiveness of the proposed method, we conduct visualization analyses on the structured semantic prior and the label representation in the latent feature space. 
First, we present the structured semantic prior about the label-to-label correspondence by visualizing the adjacency matrix, \ie, $\bm{A}^{*}$ in \cref{equ:prior_graph} on MS COCO. For ease of explanation, we select two categories, \ie, animal and vehicle, and investigate the label correspondence among labels associated with them. As shown in \cref{fig:vis}, the used structured semantic prior can successfully convey the similarity among labels although the CLIP is not encouraged in the contrastive learning on the text.
Second, we visualize the label features in the baseline (weights in the sigmoid layer) and our SCPNet (output of the SAM, denoted as ${\bm{z_i}}^{*}$ in \cref{eq:sampro}). We can observe that labels belonging to the same category are more well-aligned together in our SCPNet, compared with those in the baseline model. This result indicates that our SCPNet can reasonably derive more discriminative label representations due to the appliance of the structured semantic prior. 

\begin{figure}[!t] 
		\centering
	\begin{subfigure}{.49\linewidth}
		\centering
		\includegraphics[scale=0.2]{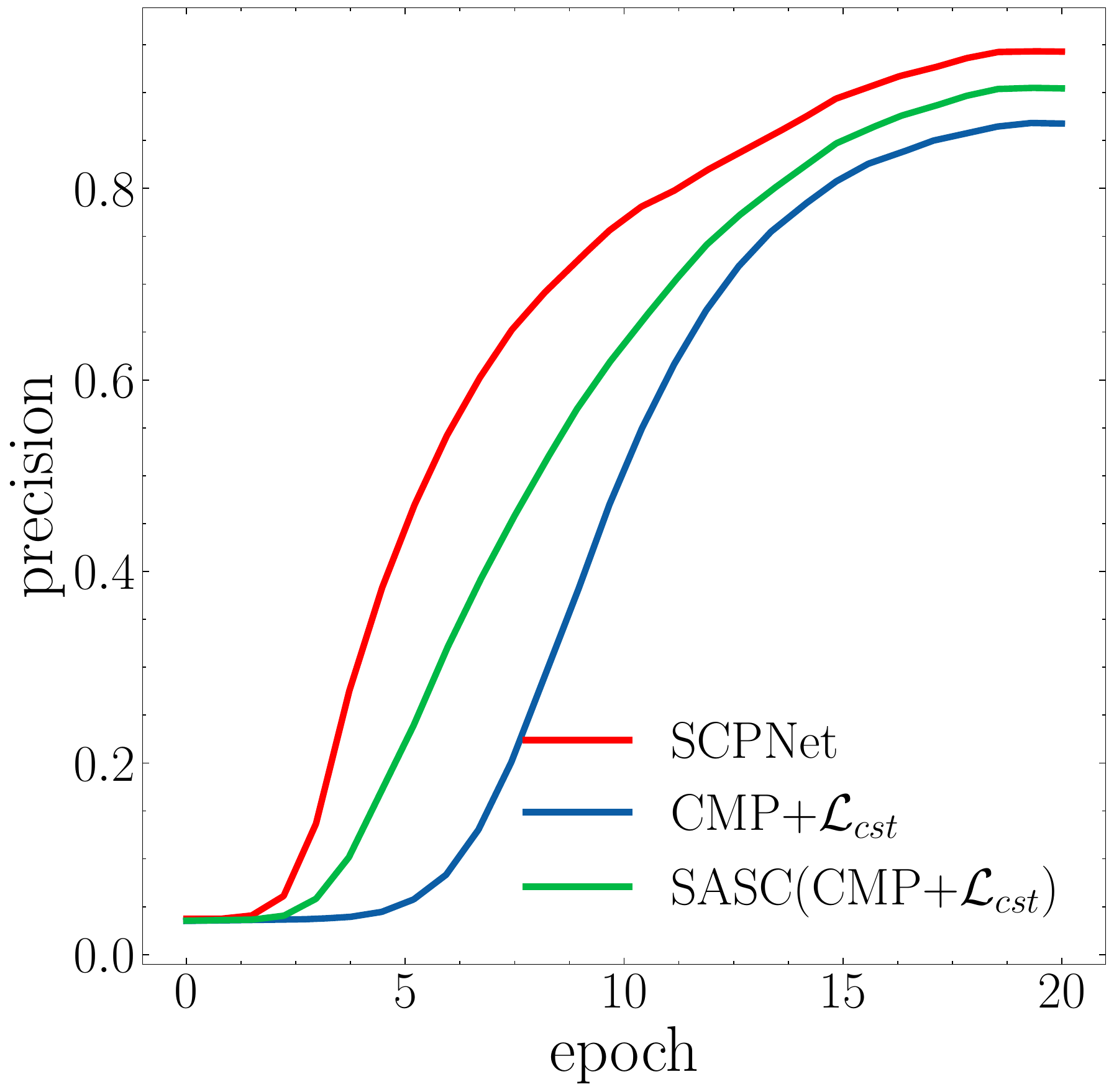}
	\end{subfigure}
	\begin{subfigure}{.49\linewidth}
		\centering
		\includegraphics[scale=0.2]{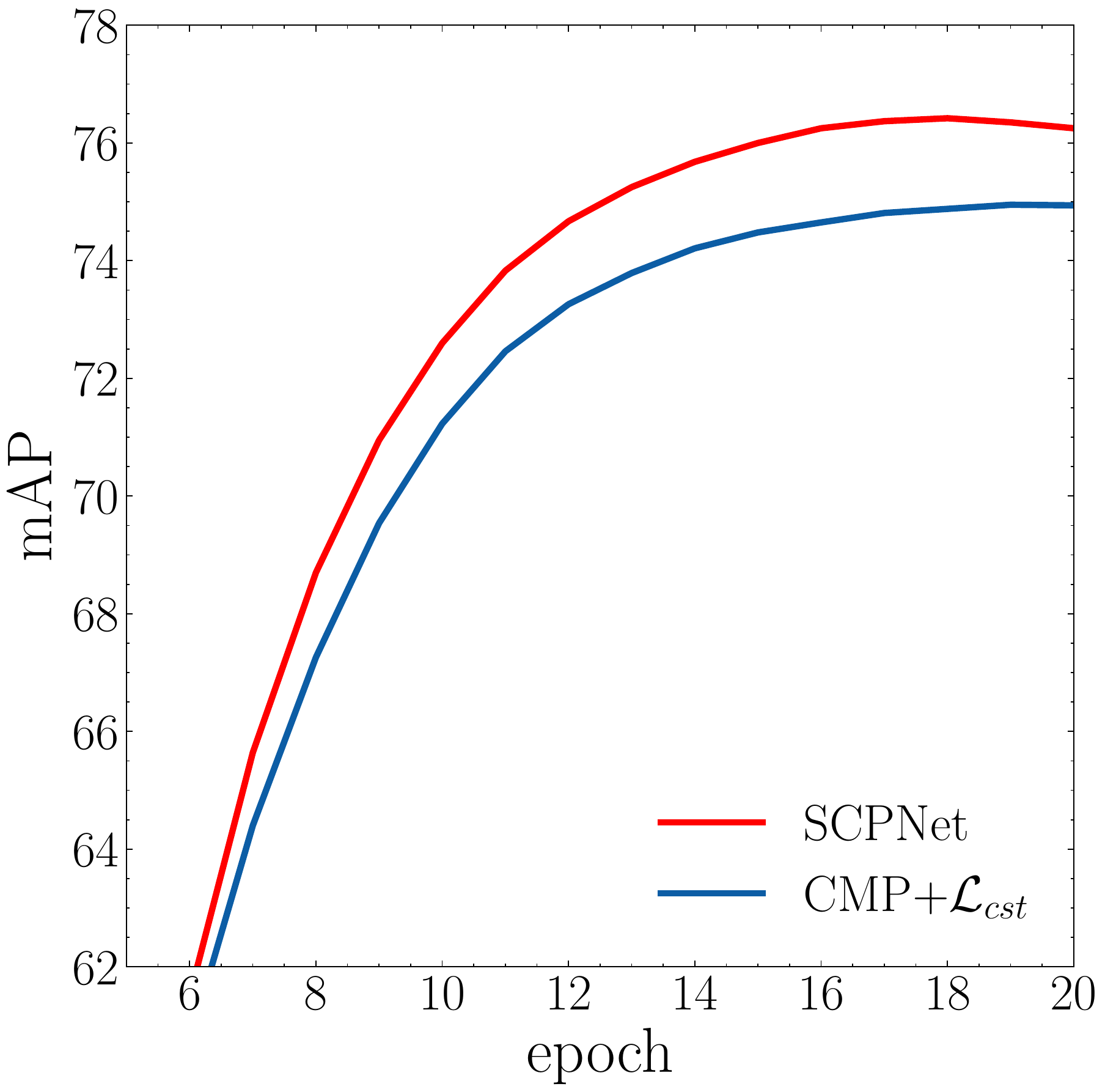}
	\end{subfigure}
	\caption{The precision on the training set (left) and the mAP on the test set (right).}
	\label{fig:precision_mAP}
\end{figure}

To verify the effect of the proposed structured semantic prior on the issue of insufficient  label supervision, we introduce a competitor model, \ie, CMP+$\mathcal{L}_{cst}$, which wipes out components involving the prior, \ie, $\bm{A}^{*}$ in \cref{equ:prior_graph}. We keep track of the precision of model predictions on the training set and the mAP result on the test set after each training epoch. For CMP+$\mathcal{L}_{cst}$, we also visualize the precision over its calibrated predictions by the $\text{SASC}(\cdot)$ function. As illustrated in \cref{fig:precision_mAP} (left), compared with CMP+$\mathcal{L}_{cst}$, both SASC$($CMP$+\mathcal{L}_{cst})$ and our SCPNet can obtain consistent improvements in terms of the label prediction precision. It indicates that the quality of label supervision can be promoted under the guidance of the proposed prior, thus benefiting the performance on the test set (see \cref{fig:precision_mAP} (right)).

\section{Conclusion}
In this paper, we drive a structured semantic prior about the label-to-label correspondence from the vision-language model, \ie, CLIP~\cite{radford2021clip}. To mitigate the deficiency of label supervision for MLR with incomplete labels, we introduce a semantic correspondence prompt network, dubbed SCPNet, which can explore such a structured semantic prior. It constructs a cross-modality prompter to leverage the explicit image-to-label correspondence in the CLIP. A semantic association module is equipped to associate related labels with the help of such a meaningful structured semantic prior. Furthermore, we propose a prior-enhanced self-supervised learning method for network optimization. Experimental results on a series of benchmark datasets for MLR with incomplete labels show that our method can achieve state-of-the-art performance on both the partial label setting and the single positive label setting, well demonstrating its effectiveness and superiority. In the future, we will further study how to generalize our method to tackle other practical problems, \eg, the domain gap.

\textbf{Acknowledgement.} This work was supported by ``Pioneer'' and ``Leading Goose'' R\&D Program of Zhejiang (No. 2023C01038), National Natural Science Foundation of China (Nos. 62271281, 61773301), Zhejiang Provincial Natural Science Foundation of China under
Grant (No. LDT23F01013F01), China Postdoctoral Science Foundation (No. BX2021161) and Shanxi Innovation Team Project (No. 2018TD-012).


{\small
\bibliographystyle{ieee_fullname}
\bibliography{egbib}

\begin{thebibliography}{10}\itemsep=-1pt

\bibitem{ben2022P-ASL}
Emanuel Ben-Baruch, Tal Ridnik, Itamar Friedman, Avi Ben-Cohen, Nadav Zamir,
  Asaf Noy, and Lihi Zelnik-Manor.
\newblock Multi-label classification with partial annotations using class-aware
  selective loss.
\newblock In {\em Proceedings of the IEEE/CVF Conference on Computer Vision and
  Pattern Recognition}, pages 4764--4772, 2022.

\bibitem{carrillo201recommender}
Dolly Carrillo, Vivian~F L{\'o}pez, and Mar{\'\i}a~N Moreno.
\newblock Multi-label classification for recommender systems.
\newblock {\em Trends in Practical Applications of Agents and Multiagent
  Systems}, pages 181--188, 2013.

\bibitem{chen2020KGGR}
Tianshui Chen, Liang Lin, Xiaolu Hui, Riquan Chen, and Hefeng Wu.
\newblock Knowledge-guided multi-label few-shot learning for general image
  recognition.
\newblock {\em IEEE Transactions on Pattern Analysis and Machine Intelligence},
  2020.

\bibitem{chen2022SST}
Tianshui Chen, Tao Pu, Hefeng Wu, Yuan Xie, and Liang Lin.
\newblock Structured semantic transfer for multi-label recognition with partial
  labels.
\newblock In {\em Proceedings of the AAAI conference on artificial
  intelligence}, volume~36, pages 339--346, 2022.

\bibitem{chen2019SSGRL}
Tianshui Chen, Muxin Xu, Xiaolu Hui, Hefeng Wu, and Liang Lin.
\newblock Learning semantic-specific graph representation for multi-label image
  recognition.
\newblock In {\em Proceedings of the IEEE/CVF international conference on
  computer vision}, pages 522--531, 2019.

\bibitem{chen2019ML-GCN}
Zhao-Min Chen, Xiu-Shen Wei, Peng Wang, and Yanwen Guo.
\newblock Multi-label image recognition with graph convolutional networks.
\newblock In {\em Proceedings of the IEEE/CVF conference on computer vision and
  pattern recognition}, pages 5177--5186, 2019.

\bibitem{chua2009nus}
Tat-Seng Chua, Jinhui Tang, Richang Hong, Haojie Li, Zhiping Luo, and Yantao
  Zheng.
\newblock Nus-wide: a real-world web image database from national university of
  singapore.
\newblock In {\em Proceedings of the ACM international conference on image and
  video retrieval}, pages 1--9, 2009.

\bibitem{cole2021multi}
Elijah Cole, Oisin Mac~Aodha, Titouan Lorieul, Pietro Perona, Dan Morris, and
  Nebojsa Jojic.
\newblock Multi-label learning from single positive labels.
\newblock In {\em Proceedings of the IEEE/CVF Conference on Computer Vision and
  Pattern Recognition}, pages 933--942, 2021.

\bibitem{durand2019learning}
Thibaut Durand, Nazanin Mehrasa, and Greg Mori.
\newblock Learning a deep convnet for multi-label classification with partial
  labels.
\newblock In {\em Proceedings of the IEEE/CVF conference on computer vision and
  pattern recognition}, pages 647--657, 2019.

\bibitem{everingham2010voc07}
Mark Everingham, Luc Van~Gool, Christopher~KI Williams, John Winn, and Andrew
  Zisserman.
\newblock The pascal visual object classes (voc) challenge.
\newblock {\em International journal of computer vision}, 88(2):303--338, 2010.

\bibitem{everingham2012voc12}
Mark Everingham and John Winn.
\newblock The pascal visual object classes challenge 2012 (voc2012) development
  kit.
\newblock {\em Pattern Anal. Stat. Model. Comput. Learn., Tech. Rep},
  2007:1--45, 2012.

\bibitem{gao2020promptfinetune}
Tianyu Gao, Adam Fisch, and Danqi Chen.
\newblock Making pre-trained language models better few-shot learners.
\newblock {\em arXiv preprint arXiv:2012.15723}, 2020.

\bibitem{huynh2020IMCL}
Dat Huynh and Ehsan Elhamifar.
\newblock Interactive multi-label cnn learning with partial labels.
\newblock In {\em Proceedings of the IEEE/CVF Conference on Computer Vision and
  Pattern Recognition}, pages 9423--9432, 2020.

\bibitem{jiang2020promptcan}
Zhengbao Jiang, Frank~F Xu, Jun Araki, and Graham Neubig.
\newblock How can we know what language models know?
\newblock {\em Transactions of the Association for Computational Linguistics},
  8:423--438, 2020.

\bibitem{kenton2019bert}
Jacob Devlin Ming-Wei~Chang Kenton and Lee~Kristina Toutanova.
\newblock Bert: Pre-training of deep bidirectional transformers for language
  understanding.
\newblock In {\em Proceedings of NAACL-HLT}, pages 4171--4186, 2019.

\bibitem{kim2022large}
Youngwook Kim, Jae~Myung Kim, Zeynep Akata, and Jungwoo Lee.
\newblock Large loss matters in weakly supervised multi-label classification.
\newblock In {\em Proceedings of the IEEE/CVF Conference on Computer Vision and
  Pattern Recognition}, pages 14156--14165, 2022.

\bibitem{krasin2017openimages}
Ivan Krasin, Tom Duerig, Neil Alldrin, Vittorio Ferrari, Sami Abu-El-Haija,
  Alina Kuznetsova, Hassan Rom, Jasper Uijlings, Stefan Popov, Andreas Veit,
  et~al.
\newblock Openimages: A public dataset for large-scale multi-label and
  multi-class image classification.
\newblock {\em Dataset available from https://github. com/openimages}, 2(3):18,
  2017.

\bibitem{krishna2017VG}
Ranjay Krishna, Yuke Zhu, Oliver Groth, Justin Johnson, Kenji Hata, Joshua
  Kravitz, Stephanie Chen, Yannis Kalantidis, Li-Jia Li, David~A Shamma, et~al.
\newblock Visual genome: Connecting language and vision using crowdsourced
  dense image annotations.
\newblock {\em International journal of computer vision}, 123(1):32--73, 2017.

\bibitem{lin2014microsoft}
Tsung-Yi Lin, Michael Maire, Serge Belongie, James Hays, Pietro Perona, Deva
  Ramanan, Piotr Doll{\'a}r, and C~Lawrence Zitnick.
\newblock Microsoft coco: Common objects in context.
\newblock In {\em European conference on computer vision}, pages 740--755.
  Springer, 2014.

\bibitem{pennington2014glove}
Jeffrey Pennington, Richard Socher, and Christopher Manning.
\newblock Glove: Global vectors for word representation.
\newblock In {\em EMNLP}, pages 1532--1543, 2014.

\bibitem{pu2022SARB}
Tao Pu, Tianshui Chen, Hefeng Wu, and Liang Lin.
\newblock Semantic-aware representation blending for multi-label image
  recognition with partial labels.
\newblock {\em arXiv preprint arXiv:2203.02172}, 2022.

\bibitem{radford2021clip}
Alec Radford, Jong~Wook Kim, Chris Hallacy, Aditya Ramesh, Gabriel Goh,
  Sandhini Agarwal, Girish Sastry, Amanda Askell, Pamela Mishkin, Jack Clark,
  et~al.
\newblock Learning transferable visual models from natural language
  supervision.
\newblock In {\em International Conference on Machine Learning}, pages
  8748--8763. PMLR, 2021.

\bibitem{shin2020autoprompt}
Taylor Shin, Yasaman Razeghi, Robert~L Logan~IV, Eric Wallace, and Sameer
  Singh.
\newblock Autoprompt: Eliciting knowledge from language models with
  automatically generated prompts.
\newblock {\em arXiv preprint arXiv:2010.15980}, 2020.

\bibitem{sivic2006videosearch}
Josef Sivic and Andrew Zisserman.
\newblock Video google: Efficient visual search of videos.
\newblock In {\em Toward category-level object recognition}, pages 127--144.
  Springer, 2006.

\bibitem{sohn2020fixmatch}
Kihyuk Sohn, David Berthelot, Nicholas Carlini, Zizhao Zhang, Han Zhang,
  Colin~A Raffel, Ekin~Dogus Cubuk, Alexey Kurakin, and Chun-Liang Li.
\newblock Fixmatch: Simplifying semi-supervised learning with consistency and
  confidence.
\newblock {\em Advances in neural information processing systems}, 33:596--608,
  2020.

\bibitem{sun2022dualcoop}
Ximeng Sun, Ping Hu, and Kate Saenko.
\newblock Dualcoop: Fast adaptation to multi-label recognition with limited
  annotations.
\newblock {\em arXiv preprint arXiv:2206.09541}, 2022.

\bibitem{tautkute2019searchengine}
Ivona Tautkute, Tomasz Trzci{\'n}ski, Aleksander~P Skorupa, {\L}ukasz Brocki,
  and Krzysztof Marasek.
\newblock Deepstyle: Multimodal search engine for fashion and interior design.
\newblock {\em IEEE Access}, 7:84613--84628, 2019.

\bibitem{wah2011CUB}
Catherine Wah, Steve Branson, Peter Welinder, Pietro Perona, and Serge
  Belongie.
\newblock The caltech-ucsd birds-200-2011 dataset.
\newblock 2011.

\bibitem{wang2020KSSNet}
Ya Wang, Dongliang He, Fu Li, Xiang Long, Zhichao Zhou, Jinwen Ma, and Shilei
  Wen.
\newblock Multi-label classification with label graph superimposing.
\newblock In {\em Proceedings of the AAAI Conference on Artificial
  Intelligence}, volume~34, pages 12265--12272, 2020.

\bibitem{yazici2020orderless}
Vacit~Oguz Yazici, Abel Gonzalez-Garcia, Arnau Ramisa, Bartlomiej Twardowski,
  and Joost van~de Weijer.
\newblock Orderless recurrent models for multi-label classification.
\newblock In {\em Proceedings of the IEEE/CVF Conference on Computer Vision and
  Pattern Recognition}, pages 13440--13449, 2020.

\bibitem{zhang2021flexmatch}
Bowen Zhang, Yidong Wang, Wenxin Hou, Hao Wu, Jindong Wang, Manabu Okumura, and
  Takahiro Shinozaki.
\newblock Flexmatch: Boosting semi-supervised learning with curriculum pseudo
  labeling.
\newblock {\em Advances in Neural Information Processing Systems},
  34:18408--18419, 2021.

\bibitem{zhang2021simple}
Youcai Zhang, Yuhao Cheng, Xinyu Huang, Fei Wen, Rui Feng, Yaqian Li, and
  Yandong Guo.
\newblock Simple and robust loss design for multi-label learning with missing
  labels.
\newblock {\em arXiv preprint arXiv:2112.07368}, 2021.

\bibitem{zheng2014contextrecommendation}
Yong Zheng, Bamshad Mobasher, and Robin Burke.
\newblock Context recommendation using multi-label classification.
\newblock In {\em 2014 IEEE/WIC/ACM International Joint Conferences on Web
  Intelligence (WI) and Intelligent Agent Technologies (IAT)}, volume~2, pages
  288--295. IEEE, 2014.

\bibitem{zhou2022cocoop}
Kaiyang Zhou, Jingkang Yang, Chen~Change Loy, and Ziwei Liu.
\newblock Conditional prompt learning for vision-language models.
\newblock In {\em Proceedings of the IEEE/CVF Conference on Computer Vision and
  Pattern Recognition}, pages 16816--16825, 2022.

\bibitem{zhou2022coop}
Kaiyang Zhou, Jingkang Yang, Chen~Change Loy, and Ziwei Liu.
\newblock Learning to prompt for vision-language models.
\newblock {\em International Journal of Computer Vision}, 130(9):2337--2348,
  2022.

\end{thebibliography}
}

\clearpage

\twocolumn[
\begin{@twocolumnfalse}
\centering
\section*{\Large Exploring Structured Semantic Prior \\ for Multi Label Recognition with Incomplete Labels \\{\large Supplementary Material}}
\end{@twocolumnfalse}
]

\appendix

\section{Methodology}
\textbf{Multi-label classification objective.}
We use the SPLC + Focal margin\cite{zhang2021simple} loss to optimize the whole model. For an input image, we denote its visual representation as $\bm{f}$ and the label feature as $\bm{{z}_i}$, which can be derived by the image encoder and text encoder, respectively (see Eq.\,({\color{red}5})). The multi-classification loss $\mathcal{L}_{cls}$ can be computed as
\begin{equation}
	\label{eq:splc}
	\begin{aligned}
		{\mathcal{L}_{cls}} = - \sum\limits_{c = 1}^C & \Big \{ {y_c}{{(1 - p_c^m)}^\alpha }\log (p_c^m) + (1 - {y_c}) \\[1mm]
		& \big [ \mathbb{I}(p \le \beta ){p_c}^\alpha \log (1 - {p_c}) + \\[3mm]
		& (1 - \mathbb{I}(p \le \beta )) {{(1 - {p_c})}^\alpha } \log ({p_c}) \big ] \Big \},
	\end{aligned}
\end{equation}
where $ p_c^m = \sigma (\text{sim} (\bm{f},\bm{{z_i}})/\tau - m)$ is the likelihood and $m$ is a margin parameter. $\alpha$ is set to 2 and $\beta$ is a threshold to identify negative label.

\section{Experiment Settings}
\subsection{Datesets for MLR with incomplete labels}
For the single positive label setting, we conduct experiments on four standard benchmarks, \ie, MS-COCO (COCO), PASCAL VOC 2012 (VOC), NUSWIDE (NUS) and CUB. The statistics of all benchmark datasets on training datasets are shown in \cref{tab:dataset}. COCO contains 82,081 training images with 80 classes and a test set of 40,137 images. VOC consists of 5,717 training images with 20 classes and 5,823 images for test. NUS is a public multi-label image classification dataset which contains 269,648 images and each image is manually annotated with some of 81 categories. CUB consists of 5,994 training images covering 312 categories and 5,794 test images. For a fair comparison with ~\cite{kim2022large},~\cite{zhang2021simple}, we perform two different setups. The LargeLoss setup divides the training dataset into 80$\%$ for training and 20$\%$ for validation. The SPLC setup only trains on the overall training set and tests on the test set. The validation sets and test sets are always fully labeled. 

For the partial label setting, we adopt three benchmarks, \ie, MS-COCO (COCO), PASCAL VOC 2007 (VOC2007) and Visual Genome (VG-200).
COCO dataset is same as the one used in the single positive label setting.
VOC2007 contains a training set of 5,011 images and a test set of 4,952 images.
VG-200 contains a total of 108,249 images covering tens of thousands of classes, most of which have only few samples. Following \cite{pu2022SARB}, we choose 200 frequent classes as the VG-200 subset, in which 10,000 images are randomly selected as the test set and the remaining 82,904 images are used as the training set.

\begin{table*}[]
	\centering
	\caption{The statistics of all benchmark datasets on training sets.}
	\renewcommand{\arraystretch}{1.2}
		\begin{tabular}{c|c|cccc}
			\hline
			Experiment setting                     & Dataset         & Samples   & Classes & Labels     & Avg.label/img \\ \hline
			\multirow{5}{*}{Single positive label} & COCO            & 82,081    & 80      & 241,035    & 2.9           \\
			& VOC             & 5,717     & 20      & 8,331      & 1.5           \\
			& NUS (LargeLoss) & 150,000   & 81      & 284,611    & 1.9           \\
			& NUS (SPLC)      & 119,103   & 81      & 289,460    & 2.4           \\
			& CUB             & 5,994     & 312     & 188,343    & 31.4          \\ \hline
			\multirow{3}{*}{Partial labels}        & COCO            & 82,081    & 80      & 241,035    & 2.9           \\
			& VOC2007         & 5,011     & 20      & 7,306      & 1.5           \\
			& VG-200          & 82,904    & 200     & 886,618    & 10.7          \\ \hline
			Real partial labels                    & OpenImages V3   & 3,552,103 & 5,000   & 13,440,371 & 3.8           \\ \hline
		\end{tabular}
	\label{tab:dataset}
\end{table*}
	
\subsection{Implementation details}
For the single positive label setting, we use a single GPU with batch size 128. Each image of ours is uniformly resized to 224 × 224 while other methods resize to 448 × 448. We use the Adam optimizer and OneCycle learning rate schedule with the max learning rate of 3e-5. It is trained with 30 epochs in total.

For the partial label setting, we use two GPUs with batch size 32 and the max learning rate is 1e-5. We also train for 30 epochs on all benchmark datasets. 
	
For data augmentation, we adopt the random horizontal flip and random resized crop for the weak transformation and the RandAugment for the strong transformation.
	
\section{More experiments results}
\subsection{Model Analysis}
\textbf{Effect of different modules on the ResNet.}
To investigate the effectiveness of our proposed method with CNN-based MLR models, we conduct the ablation study on replacing the pretrained CLIP with the ResNet model. As shown in \cref{tab:resnet_ablation}, each component can lead to performance improvement. Compared with the baseline, introducing SAM can achieve a performance improvement of $1.01\%$, which shows that exploiting the implicit label-to-label correspondence can benefit the MLR with incomplete labels. In addition, using the overall proposed PESSL can accomplish $0.53\%$ mAP improvement, well demonstrating the advantage of incorporating the structured semantic prior to calibrate the semantic distribution. Finally, our proposed method significantly outperforms the baseline model with $1.54\%$ improvement, well indicating the effectiveness and superiority of our method.

\begin{table}[]
	\centering
	\caption{Analysis of different modules on the ResNet50 (\%).}
	\renewcommand{\arraystretch}{1.2}
	\begin{tabular}{c|cc|c}
		\hline
		ResNet      & SAM         & PESSL        & mAP   \\ \hline
		\checkmark  &             &              & 73.18 \\ 
		\checkmark  & \checkmark  &              & 74.19 \\
		\checkmark  & \checkmark  & \checkmark   & \textbf{74.72} \\ \hline
	\end{tabular}
	\label{tab:resnet_ablation}
\end{table}

\begin{table}[]
	\centering
	\renewcommand{\arraystretch}{1.2}
	\caption{Analysis on different combination of encoders (\%). ResNet50 is pretrained on the ImageNet. ResNet50$_r$ is randomly initialized.}
	\begin{tabular}{c|c|c}
		\hline
		Image Encoder   & Label Encoder  & mAP    \\ \hline
		ResNet50        & Linear         & 73.18       \\ 
		CLIP            & Linear         & 73.33       \\ \hline
		ResNet50        & CLIP           & 64.85       \\
		ResNet50$_r$    & CLIP           & 41.71       \\
		CLIP            & CLIP           & \textbf{74.36}  \\ 
		\hline 
	\end{tabular}
	\label{tab:cmp_encoder}
\end{table}

\textbf{Combination of different encoders.} We investigate different combinations of image encoders and label encoders. As shown in \cref{tab:cmp_encoder}, compared with our method (the last row), replacing the CLIP label encoder with a linear prediction layer leads to inferior performance for a pretrained/CLIP-based ResNet50 (Row 1/2). Besides, replacing the CLIP image encoder with a pretrained/random ResNet50 (Row 3/4) also encounters great performance drops due to the destruction of image-label correspondence. This evidence shows the superiority of applying CLIP as the base MLR model.

\textbf{Effect of the prompt learning.} We investigate the effect of the prompt learning with the CMP model. As shown in \cref{tab:cmp_prompt}, we can observe that with the random initialization, different prompt length can obtain similar performance. After leveraging the prompt template, \ie, \textit{a photo of a}, CMP with 4 prompts can achieve the best performance, indicating the positive effect of the hard prompt~\cite{radford2021clip} as the prompt initialization.

\begin{table}[]
	\centering
	\caption{Effect of the prompt in the CMP (\%).}
	\renewcommand{\arraystretch}{1.2}
	\begin{tabular}{c|c|c}
		\hline
		Length        & Initialization        & mAP   \\ \hline
		4             & template              & \textbf{74.36} \\
		4             & Random                & 73.85 \\
		8             & Random                & 73.87 \\
		16            & Random                & 73.85 \\ \hline
	\end{tabular}
	\label{tab:cmp_prompt}
\end{table}

\begin{table}[]
	\centering
	\caption{Analysis on the semantic association module (SAM) (\%).}
	\renewcommand{\arraystretch}{1.2}
	\begin{tabular}{c|c|c|c}
		\hline
		\multirow{2}{*}{Label Feature} & $\bm{H}^L$  & $\bm{H}^0, \bm{H}^L$  & $\bm{H}^0 + \bm{H}^L$  \\ \cline{2-4}
		& 75.01       & 75.65      & \textbf{76.42} \\ \hline
		\multirow{2}{*}{Correlation Matrix} & Original    & Sparse   & Ours  \\ \cline{2-4}
		& 75.96       &76.13       &\textbf{76.42}  \\ \hline
	\end{tabular}
	\label{tab:SAM}
\end{table}

\textbf{Analysis on SAM.} For the proposed SAM component, we first investigate how to construct the label feature. We discuss three variants: 1) directly using $\bm{H}^L$, 2) using $\bm{H}^0$ and $\bm{H}^L$ in a multi-task learning strategy, and 3) the proposed residual connection, \ie, $\bm{H}^0+\bm{H}^L$. From \cref{tab:SAM}, we can see that the residual connection obtains the best performance. We further analyze the effect of different label correlation matrix. We verify three strategies: 1) using original label correlation matrix (see Eq.\,({\color{red}1})), 2) resulting in a sparse matrix by retaining the top $K$ elements (see Eq.\,({\color{red}2})), and 3) our adjusted the sparse matrix (see Eq.\,({\color{red}4})), which is our proposed structured semantic prior, \ie, $\bm{A^*}$. As shown in \cref{tab:SAM}, our proposed method can achieve the state-of-the-art performance, demonstrating that the label-to-label correspondence can be well captured by the proposed structured semantic prior.

\textbf{Analysis of the pseudo label selection.} 
We further investigate different pseudo label selection for weak transformation in the PESSL. We discuss three strategies: 1) using the weak transformation prediction probabilities as soft label, 2) setting threshold to filter the pseudo label, and 3) our method which further selects the top highest probability to construct a set of confident labels. As shown in \cref{tab:pseudo label}, our method obtains the best performance, demonstrating that the selective construction of confidence labels is more compatible with the MLR task.

\begin{figure*}
	\includegraphics[width=\linewidth]{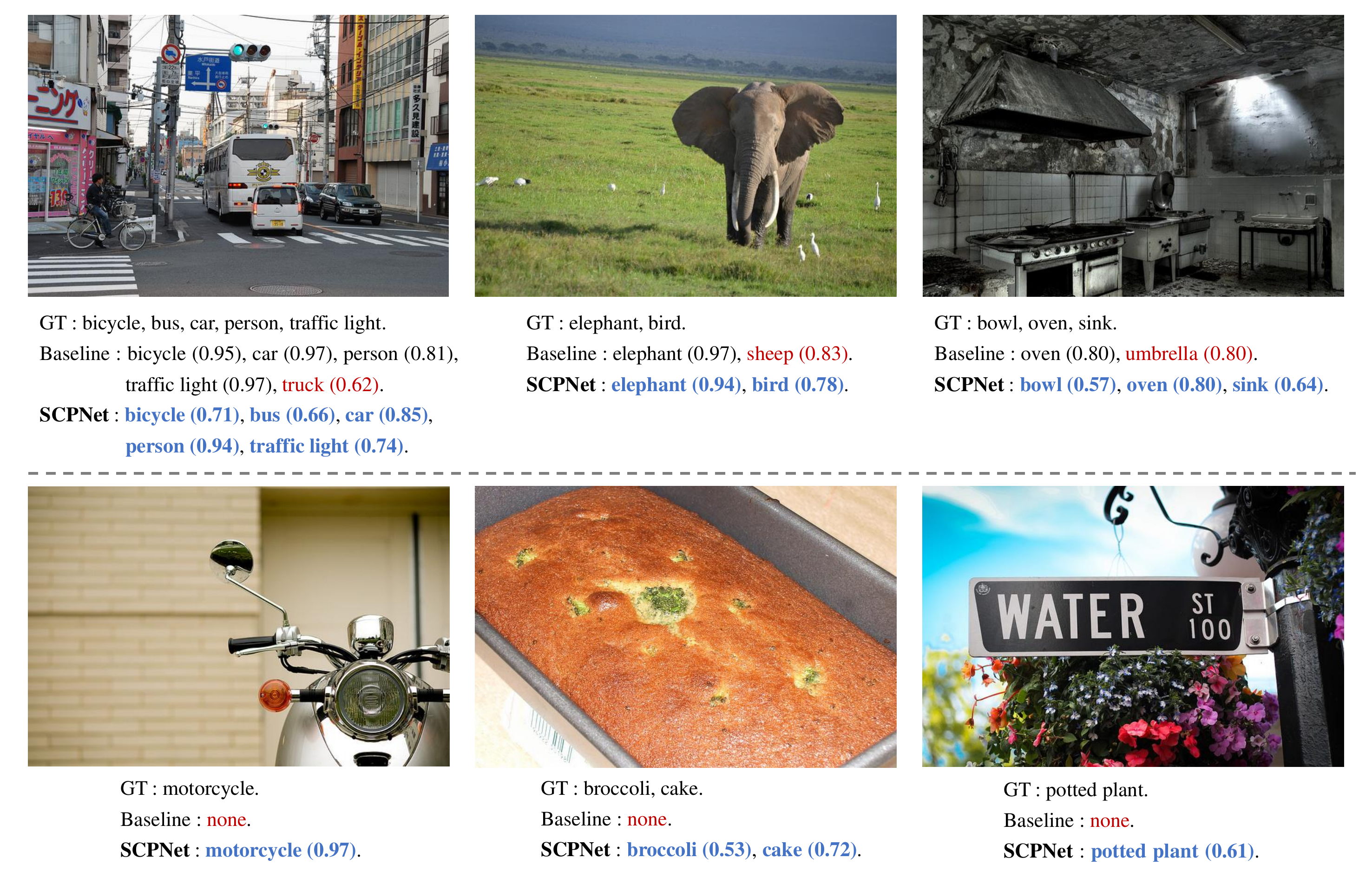}
	\caption{Visualization of example results compared our SCPNet with the baseline model. Red color means results of false recognition and missing recognition. Blue color denotes ours results. GT means ground truth and the precision of recognition is in brackets.}
	\label{fig:results}
\end{figure*}

\begin{table}[]
	\centering
	\caption{Different pseudo label selection of PESSL (\%).}
	\renewcommand{\arraystretch}{1.2}
	\begin{tabular}{c|c|c|c}
		\hline
		$\mathcal{L}_{cst}$  & Soft Label  & Threshold    & Ours      \\ \hline
		mAP                  & 75.88   & 75.80   & \textbf{76.42} \\ \hline
	\end{tabular}
	\label{tab:pseudo label}
\end{table}

\textbf{Stress-testing on domain-specific datasets.}
We conduct experiments on a common satellite dataset (AID\footnote{https://github.com/Hua-YS/AID-Multilabel-Dataset}) and a common medical dataset (ChestX-ray14\footnote{https://nihcc.app.box.com/v/ChestXray-NIHCC}) under the single positive label setting to perform stress-testing on domain-specific datasets which are far from those used in CLIP pretraining. As shown in \cref{tab:domain specific}, compared with CMP, our SCPNet can obtain $5.84\%$ and $5.50\%$ performance improvement on the AID and ChestX-ray14, respectively. These results show that our method achieves the best performance although CLIP cannot generalize well in these datasets, well demonstrating the generalization. 

\begin{table}[]
	\centering
	\caption{Stress-testing on domain-specific datasets (\%).}
	\renewcommand{\arraystretch}{1.2}
	\begin{tabular}{c|cc}
		\hline
		Method                      & AID            & ChestX-ray14     \\ \hline
		SPLC\cite{zhang2021simple}  & 71.26          & 25.60            \\
		CMP (ours)                  & 67.48          & 22.42            \\ 
		\textbf{SCPNet (ours)}      & \textbf{73.32} & \textbf{27.92}   \\ \hline
	\end{tabular}
	\label{tab:domain specific}
\end{table}

\begin{table}[]
	\centering
	\caption{DualCoOp vs. SCPNet in terms of computation cost}
	\renewcommand{\arraystretch}{1}
	\begin{tabular}{c|cc}
		\hline
		& DualCoOp   & SCPNet   \\ \hline
		Training speed [iters/sec]    & 4.27     & 2.82   \\
		Trainable parameters          & 1.31M      & 3.41M   \\
		GPU memory for training       & 7.4G   &  9.8G \\
		Inference speed [samples/sec] & 318.59     & 322.76   \\
		GPU memory for inference      & 3.4G   & 3.4G  \\ \hline
		mAP performance on COCO & $81.9\%$ &$83.2\%$ \\\hline
	\end{tabular}
	\label{tab:computation}
\end{table}

\textbf{DualCoOp vs. SCPNet.}
First, in terms of the model performance, for fair comparison, we implemented our method with a frozen image encoder on DualCoOp's code under the same setting, achieving $83.2\%$ mAP (+$1.3\%$). We also tuned DualCoOp under the SCPNet setting, where DualCoOp achieved inferior mAP with $82.4\%$ (vs. ours: $83.8\%$).
Second, in terms of the computation efficiency, we compare our method with DualCoOp under the DualCoOp's setting. As shown in \cref{tab:computation}, during training, SCPNet consumes more resources than DualCoOp. But the required cost is not unaffordable in practice. During inference, both methods can derive label features offline. Therefore, SCPNet is comparable to DualCoOp in terms of computational cost while enjoying the superior performance. This shows SCPNet is more advanced or at least comparable when applied in practical scenarios. 
These results clearly demonstrate the effectiveness and superiority of our method, compared to DualCoOp.

\begin{figure*}
	\includegraphics[width=\linewidth]{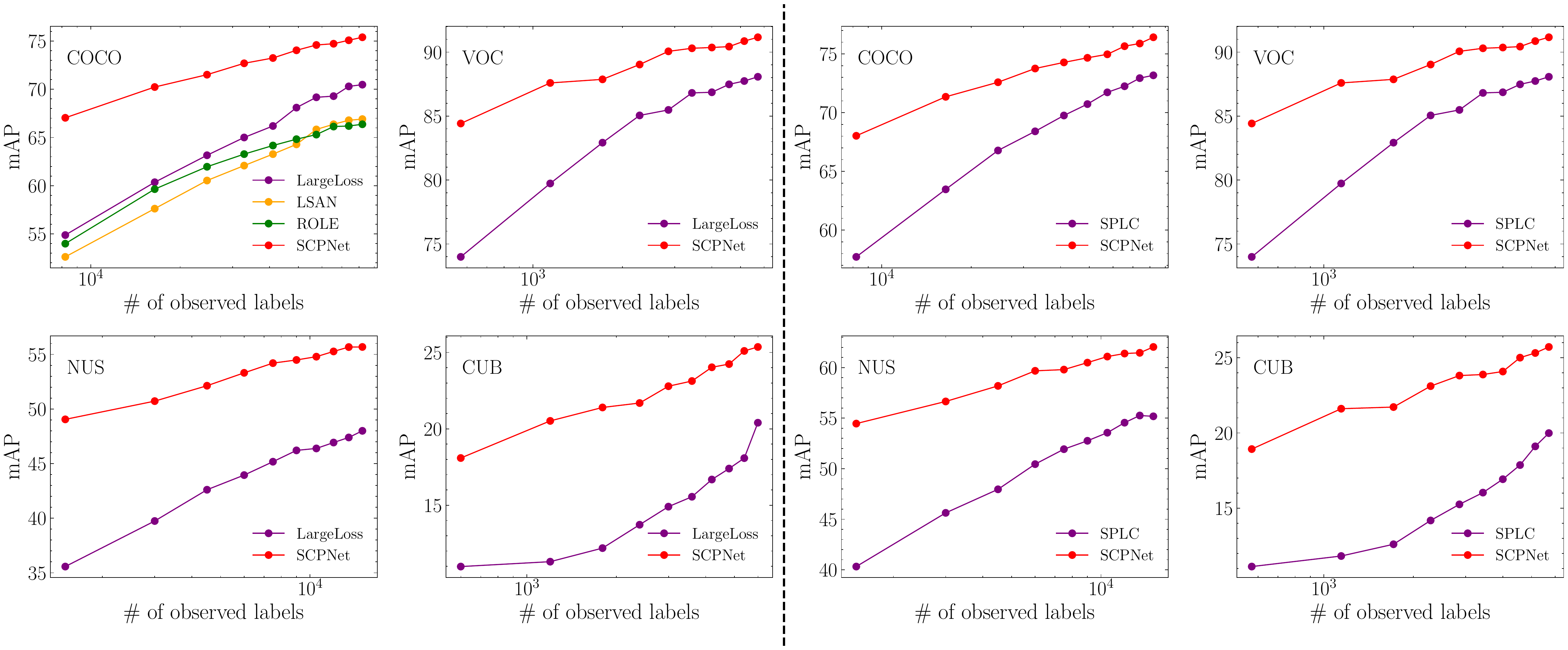}
	\caption{Results of the few-shot partial label setting on COCO, VOC, NUS and CUB dataset (left: in the LargeLoss setup, right: in the SPLC setup).}
	\label{fig:Fewshot}
\end{figure*}

\textbf{Hyper-parameters selection.}
For the single positive label setting, most hyper-parameters are directly borrowed from COCO, except for some dataset-dependent hyper-parameters, \eg, $K$ in Eq.\,({\color{red}2}) (best at $60$/$15$/$50$/$280$ for COCO/VOC/NUS/CUB). $s$ in Eq.\,({\color{red}3}) is empirically set to $0.2$. For the partial label setting, most hyper-parameters are directly borrowed from the single positive label setting, except for the learning rate. Even so, our method can obtain consistent performance improvements in all scenarios, well demonstrating the robustness of hyper-parameters. We show that after more hyperparameter searches, we can obtain slightly better performance than the reported one, \eg, $49.6\%$ vs. $49.4\%$ on VG-200.

\subsection{Multi-label Recognition Results}
Here we present the multi-label recognition results on the single positive label setting. As shown in \cref{fig:results}, our proposed method can successfully recognize more accurate labels with lower false identifications (see examples in the first row). Besides, compared with the baseline model, our method can achieve fewer missing recognition for difficult labels, \eg, ``broccoli'' in the middle of the second row. These results further demonstrate the effectiveness and the superiority of our proposed method.

\subsection{Few-Shot Single Positive Label Setting}
To investigate the effectiveness of the proposed method with a smaller number of training images, we further conduct the experiments in the few-shot single positive label setting under both the LargeLoss setup \cite{kim2022large} and the SPLC setup~\cite{zhang2021simple}.

As illustrated in \cref{fig:Fewshot} (left), in the LargeLoss setup, following \cite{kim2022large}, we randomly sample the training images from 10$\%$ to 100$\%$ and conduct experiments on the COCO dataset. We further compare our method with LargeLoss~\cite{kim2022large} on the other benchmark datasets, \ie, VOC, NUS and CUB. Only given 10$\%$ of the training images, our method can obtain a maximal performance improvement of $12.16\%$, $7.47\%$, and $13.48\%$ on COCO, VOC and NUS dataset, respectively. For CUB dataset, we achieve a maximal improvement of $9.22\%$ with giving the training images of 20$\%$. Overall, our method can accomplish an average performance improvement of $7.16\%$, $3.20\%$, $9.35\%$, and $7.62\%$ on four datasets with the training images from 10$\%$ to 100$\%$. Besides, as shown in \cref{fig:Fewshot} (right), in the SPLC setup, we present the comparison results with SPLC~\cite{zhang2021simple} on four benchmark datasets as well. The maximum performance improvement achieved by our method can reach $10.31\%$, $10.43\%$ and $14.14\%$ on COCO, VOC and NUS dateset, respectively. Our method can bring a maximum improvement of $9.78\%$ with 20$\%$ training images on CUB dataset. Our SCPNet can obtain $5.06\%$, $4.79\%$, $8.76\%$, and $7.82\%$ improvement on average for the four datasets, respectively. 

These experimental results show that our proposed SCPNet can significantly achieve state-of-the-art performance in different few-shot single positive label setting, well indicating the generalization and superiority.
\subsection{Real Partial Label Scenario}
\textbf{Dataset and implementation details.}
To analyze the effectiveness of the proposed method in real partial label scenario, we conduct experiments on the OpenImage V3\cite{krasin2017openimages} dataset with 5,000 classes. The details are shown in \cref{tab:dataset}, OpenImage V3 contains 3.5M training images, 42k validation images, and 125k test images. Follwing \cite{kim2022large}, we divide the training images into 5 groups, where G1 has the smallest number of the counted images and G5 is the largest one. All Gs corresponds to the set of all categories.

\begin{table}[]
	\centering
	\caption{Results with real partial label on OpenImage V3 dataset.}
	\renewcommand{\arraystretch}{1.2}
	\resizebox{\linewidth}{!}{
		\begin{tabular}{c|ccccc|c}
			\hline
			Method                      & G1            & G2            & G3            & G4            & G5            & All Gs         \\ \hline
			CL\cite{durand2019learning} & 70.4          & 71.3          & 76.2          & 80.5          & 86.8          & 77.1          \\		
			IMCL\cite{huynh2020IMCL}    & 71.0          & 72.6          & 77.6          & 81.8          & 87.3          & 78.1          \\
			Naive AN                    & 77.1          & 78.7          & 81.5          & 84.1          & 88.1          & 82.0          \\
			WAN\cite{cole2021multi}     & 71.8          & 72.8          & 76.3          & 79.7          & 84.7          & 77.0          \\
			LSAN\cite{cole2021multi}    & 68.4          & 69.3          & 73.7          & 77.9          & 85.6          & 75.0          \\
			LargeLoss\cite{kim2022large}& 77.7          & 79.3          & 82.1          & 84.7          & 89.4          & 82.6          \\
			P-ASL\cite{ben2022P-ASL}    & 73.2          & 78.6          & 85.1          & 87.7          & 90.6          & 83.0 \\
			\textbf{SCPNet (ours)}      & \textbf{79.6} & \textbf{81.8} & \textbf{85.3} & \textbf{87.9} & \textbf{92.1} & \textbf{85.3} \\ \hline
	\end{tabular}}
	\label{tab:openimages}
\end{table}

\textbf{Compared methods.}
We compare our method with Curriculumn Labeling(CL)\cite{durand2019learning}, IMCL\cite{huynh2020IMCL}, Naive AN, Weak AN (WAN)\cite{cole2021multi}, Label Smoothing with AN (LSAN)\cite{cole2021multi}, LargeLoss\cite{kim2022large} and P-ASL\cite{ben2022P-ASL}.

\textbf{Results.}
As shown in \cref{tab:openimages}, our method outperforms the state-of-the-art methods on G1, G2, G3, G4 and G5 with an improvement of $1.9\%$, $2.5\%$, $0.2\%$, $0.2\%$ and $1.5\%$, respectively. As a whole, our proposed SCPNet can accomplish a performance improvement of $2.3\%$ on all Gs, well demonstrating the effectiveness and generalization to practical scenarios.

\end{document}